\documentclass{article}
\usepackage{graphicx}

\usepackage{algorithmic}
\usepackage{amsmath}
\usepackage{color,soulutf8}
\usepackage[nolist,nohyperlinks]{acronym}
\usepackage{subfig}
\usepackage{amssymb}
\usepackage{filecontents}
\usepackage{multirow}
\usepackage{verbatim}
\usepackage{booktabs}

\usepackage{ctable}
\usepackage{array}
\usepackage{makecell}
\usepackage{tabularx}
\usepackage{relsize}
\usepackage{hyperref}
\usepackage{enumitem}
\hypersetup{
	linkcolor=blue,
	citecolor=blue;
	colorlinks=true,
}
\usepackage{footnote}

\usepackage[algoruled,boxed, lined]{algorithm2e}

\begin{document}

\title{Self-adaptive decision-making mechanisms to balance the execution of multiple tasks for a  multi-robots team}

\author{Nunzia Palmieri$^1$, Xin-She Yang$^2$, Floriano De Rango$^1$,
Amilcare Francesco Santamaria$^1$ \\[10pt]
1) Department of Computer Engineering, Modeling, Electronics and System Science, \\
University of Calabria, Italy. \\[10pt]
2) Design Engineering and Mathematics, School of Science and Technology,\\
Middlesex University London, UK}

\date{}

\maketitle

\begin{abstract}

 This work addresses the coordination problem of multiple robots with the goal of finding specific hazardous targets in an unknown area and dealing with them cooperatively. The desired behaviour for the robotic system entails multiple requirements, which may also be conflicting. The paper presents the problem as a constrained bi-objective optimization problem in which mobile robots must perform two specific tasks of exploration and at same time cooperation and coordination for disarming the hazardous targets. These objectives are opposed goals, in which one may be favored, but only at the expense of the other. Therefore, a good trade-off must be found. For this purpose, a nature-inspired approach and an analytical mathematical model to solve this problem considering a single equivalent weighted objective function are presented. The results of proposed coordination model, simulated in a two dimensional terrain, are showed  in order to assess the behaviour of the proposed solution to tackle this problem. We have analyzed the performance of the approach and the influence of the weights of the objective function under different conditions: static and dynamic. In this latter situation, the robots may fail under the stringent limited budget of energy or for hazardous events. The paper concludes with a critical discussion of the experimental results.
\end{abstract}

{\bf Keywords:} Multi-Robot Systems, Bi-criteria Optimization Model,  Nature-Inspired Algorithms,  Self-Coordination of Multiple Robots,  Hazardous Environment. \\[15pt]

\hrule
\vspace{5pt}
\noindent {\bf Citation detail:} 
Nunzia Palmieri, Xin-She Yang, Floriano De Rango, Amilcare Franscesco Santamaria, Self-adaptive decision-making mechanisms to balance the execution of multiple tasks for a multi-robot team, Neurocomputing, vol. 306, 17-36 (2018). \\

https://doi.org/10.1016/j.neucom.2018.03.038

\hrule

\section{Introduction}

Search for unknown targets and then management of the found targets can have great importance to many applications related to disaster situations like earthquakes, counter terrorist attacks, clean-up of minefield and handling of lost hazardous objects. Over the last decade, many researchers have been studying groups of robots that act as a team to accomplish certain difficult tasks in dynamic, unknown and hazardous environments. A team of robots can be assigned to a task that is too heavy and/or harmful to humans and such robots can accomplish the assigned task in a faster, cheaper, more efficient way, providing a better robustness and adaptability \cite{1,2}.

This paper studies the coordination of multiple robots that need to explore an unknown area, in order to detect and disarm cooperatively dangerous sources (e.g., land mines, hazardous chemicals), since it is either impossible or too expensive for only one robot to carry out the task individually. In these applications, robots must have the ability to distribute themselves among various locations of the area and also redistribute them at the target's  locations, in order to handle them in parallel.  Furthermore, the team is designed to complete the overall task in a cost-saving way, with an aim of the minimization of the mission time.
This issue is similar to a task allocation problem, but the main difference is that the locations of targets are not known {\it a priori} and the assignment can be dynamic as the number of targets found can be one or more at any time.

The most common approach is instantaneous task allocation, which means that the robots are instantaneously assigned to the targets that could give the maximum benefit at the moment and hence the task allocation is achieved according a greedy strategy. However the instantaneous task allocation does not take into account any future events and it may degrade the performance. A desirable feature of task allocation scheme, especially in a dynamic uncertain environment, is that the robots can  be well-prepared to react to new events that can occur.

Therefore, the main aim of this work is to propose an approach which attempts to simultaneously minimize the exploration time and the coordination/recruitment time. The goal is to assign each robot to the best task from its point of view, balancing exploration and recruitment and at the same time reacting in an efficient manner to the events that can occur anytime. In our proposed approach, the formulation of the problem is considered as a bi-objective model with constraints and the use of nature-inspired approaches. Then, a weighted objective function is proposed to balance the two goals, and specific values of the weights are investigated in order to analyze different scenarios in the proposed solution.
It is worth pointing out that we did not focus on the detailed mechanism of detection and disarmament of the targets such as the exact procedure of handling a hazardous target. Rather, we suppose that the target can be detected by proper sensors and dealing with through certain actions that can be modelled as a fixed time delay. Consequently, we are more interested in how to provide the coordinated motions in a distributed and decentralized decision mechanism when the information about the environment for each robot is only partially available or altogether absent.

However, two major problems exist: how to explore the area and make the decision on where to move effectively at a reasonable computational cost,  and how to avoid deadlock; that is, the situation where robots are waiting for a long time for the others to proceed to disarming process. These issues are particularly relevant in strictly collaborative tasks since the robots need to work collectively and adaptively for the disarmament of the hazardous targets, and each robot has only locally and partially information about the environment.

The paper presents a new swarm intelligence (SI)-based approach that is strongly inspired by the biological behaviour of social insects. Biology-inspired metaheuristic algorithms have recently become the forefront of the current research as an efficient way to deal with many NP-hard combinatorial optimization problems and non-linear optimization constrained problems in general \cite{3}. These algorithms are based on some particular successful mechanisms of  biological phenomena of mother nature in order to achieve the survival of the fittest in a dynamically changing environment.  Examples of collective behaviours in nature are numerous. They are based, mainly, on direct or indirect exchange of information about the environment between the members of the swarm. Although the rules governing the interactions at the local level are usually easy to describe, the result of such behaviour is difficult to predict. However, through collaboration, swarms in nature are able to solve complex problems that are crucial for their survival.

More specifically in this paper, our work combines two bio-inspired algorithms using only local interactions between the members of the swarm and with the environment. Therefore, in our approach, each robot is able to self-organize and uses only simple information. Using this information, it assesses the task opportunities individually, and makes movements through a incremental phase. The action mechanism relies on two types of communication: indirect  communication and direct communication.

First, we introduce an algorithm for the exploration of the area based on a repulsive pheromone mechanism where no Wi-Fi communication is possible. An indirect communication through depositing and sensing chemical pheromone on the grid cells is available as an exploration mechanism for the deployment of the robots in the environment.  This form of communication is embedded in the environment and it is inspired by ants that lay pheromone as marking to be sensed later by others (including possibly themselves). The environment allows the aggregation of the local interactions of numerous robots in order to achieve a certain aim. The environment becomes a shared memory on which information can be stored and removed (according to certain natural mechanisms). The algorithm is inspired by the  cooperative behaviour in nature, as the colonies of social insects \cite{4}.

On the other hand, when a robot detects a target, in order to handle it cooperatively and more quickly, it manages and drives the coordination process using direct communication. For this purpose, a modified Firefly-based algorithm is proposed as a decision and recruitment mechanism \cite{5}.

The above mentioned theoretic approach has been tested in our Java-based simulator, and it has been analyzed in terms of different values of the weights of the objective function, in order to evaluate their influence on balancing recruitment and exploration tasks and how they affect the behaviour of the swarm to stick together and/or move according a common goal.
The experiments have been repeated under different conditions such as the variations of the number of targets, the number of required robots to disarm a target, the dimension of the  area. In addition, the paper also has explored how the coefficient may influence the performance both in static and dynamic scenarios.

Therefore, the remainder of the paper is organized as follows. Section 2 provides a review of the related work. The descriptions of the problem and the mathematical model are the focus of Section 3 and Section 4. Section 5 describes the proposed bio-inspired approach. Section 6 shows the simulation results obtained from a set of experiments and finally Section 7 outlines the main research conclusions.

\section{Related work}

\subsection{Multi-robot Allocation and Coordination}

The applications of multi-objective optimization to evolutionary robotics are diverse
with increasing attention. Some common applications for multi-robot teams include foraging \cite{6}, path planning \cite{7}, search \cite {8}, distributed surveillance and security \cite{9}, search and rescue \cite{10}\cite{11}, and emergency service \cite{12,13,14}.  However, many of these formulations have attempted to solve multiple objectives at the same time and researchers often model these common capabilities by defining them as optimization problems. Typically, these problems can be formulated as combinatorial optimization or as convex optimization problems in special cases so as to take advantage of the many tools available for these type of optimization.
However, such formulations were often simplified and were not treated as global optimization problems for multi-robot applications, though the full combinatorial problems related to robot exploration and coordinations can be NP-hard. Since most robotic applications require real-time robot responses, there is insufficient time to calculate globally optimal solutions for most applications; such solutions are generally possible for very small-scale problems. Instead, typical solutions use distributed methods that incorporate only local cost/utility metrics. Although such approaches can only achieve approximations to the global solution, they often can be sufficient for practical applications \cite{15}.

On the other hand, a diverse range of studies has been done by imitating ideas from nature for designing control algorithms for multi-robot systems, since the nature presents excellent examples of distributed self-organizing systems. The aim is to develop self adaptive distributed coordination algorithms to deal with the problems. In nature, we often see complex group behaviors arisen from biological systems composed by large numbers of animals that individually lack either the communication and computational capabilities, but they may self-organize, leading to some emerging collective behavior that enable to achieve a common (swarm level) goal \cite{16}.

Regarding the exploration  task, the main goal is to cover the whole area in the minimum possible time. Therefore, it is essential that the robots are deployed in different regions of the area, at the same time. Most researches have investigated the use of indirect communication as a mechanism to guide the swarm of the robots. Stigmergy is a kind of mechanism that mediates animal-animal interactions through artifacts or via indirect communication, providing a kind of environmental synergy, information gathered from work in progress, distributed incremental learning and memory among the society.
Furthermore, pheromone provides a stigmergic medium of communication, which influences the future actions of a single or a group of individuals via the changes made to the environment. Stigmergy allows to record the history of an agent’s actions without memorizing the information and saving resources \cite{17}.

In the exploration task, researchers use the concept of anti-pheromone so as to try to maximize the distance between the robots and to enforce a dispersion mechanism in different sites of the region of interest, with the aim to accomplish the mission as quickly as possible. Some examples of this approach can be found in  \cite{18,19} for surveillance mission, in \cite{20} for guiding the robots in search and rescue in a disaster site, in \cite{21,22} in multi-robot coverage.  Ravankar et al. \cite{23} used a hybrid communication framework that incorporates the repelling behaviour of the anti-pheromone and attractive behaviour of pheromone for efficient map exploration.

Regarding the coordination of the robot for task assignment or allocation, many approaches have been proposed for solving the multi-robot coordination problem.  One of the most commonly used swarm-based approaches is the response threshold, where each robot has a stimulus associated with each task it has to execute. Some response threshold systems use such stimuli and the threshold value for calculating the probability of executing a task \cite{24,25}.
In recent years, market-based approaches have become popular to coordinate multi-robot systems. These methods have attempted to present a distributed solution for the task allocation problem \cite{26}. Jones et al. \cite{27} described a market based approach to task allocation for the fire fighting in a disaster response domain.

Recently, bio-inspired algorithms inspired by a variety of biological systems, have been proposed for self-organized robots. A well known bio-inspired approach takes inspiration from the behaviour of the birds, called Particle Swarm Optimization (PSO). PSO-inspired methods and their extended versions have received much attention and have been applied for the coordination of mobile robots. Some examples of its application can be found  for guiding robots for targets searching in complex and noisy environment as presented in \cite{28}. Modified versions of the PSO are proposed to balance searching and selecting in a collective clean-up task \cite{29} for path planning in a clutter environment \cite{47} and for mimicking natural selection emulated using the principles of social exclusion and inclusion \cite{30}.

Other studies took inspiration from the bees and ants that mimic the food foraging behaviour of swarms of honey bees and ants in nature. These algorithms have also been applied to robotic systems such as task allocation \cite{31}, finding targets and avoiding obstacles \cite{32}, for solving on line path planning \cite{33} \cite{34}, decision-making to aggregate robots around a zone  \cite{13} \cite{16} \cite{35}. A Hybrid approach can be found in \cite{36}. However, extensive reviews of research related to the bio-inspired techniques and the most behaviour of the robots can be found in \cite{37} \cite{38}.

\subsection{Overview of the Present Paper }

As discussed above, the problem of multi-robot coordination has received significant attention. However, the proper coordination for exploration and management of found targets have not been studied extensively. Therefore, the research we report here makes some new contributions in this area:
\begin{itemize}
\item The exploration and handling problem is described using an optimization model.
 Both the search and handling of the hazard targets are considered together, allowing the trade-off between two aims in terms of a weighted objective function.

\item A repulsive mechanism based on ant colony optimization is applied as an indirect coordination mechanism for the exploration task.

\item The recruitment task that allows the disarming of a hazardous target cooperatively is modelled using  a new bio-inspired Firefly Algorithm based approach to recruit and then coordinate robots movements as a decision mechanism of the robots.

\item The environment is highly uncertain where no prior information is available and dynamic events such as robots failures under a limited amount of battery power and  unpredictably events in the area can occur.

\item The robots react to events that occur in the environment, so unlike other common approaches, they could change, at each time step, the previous decisions taken earlier.
\end{itemize}

 With these more realistic considerations, we present a hybrid bio-inspired approach that allows robots to potentially balance the needs of search and task response in an adaptive way to complete the mission, obtaining full benefits in terms of reducing wastage of resources of the system such as time, energy and mobile alive robots.

\section{Scenario and Model Formulation}

In this paper we address challenges in twofold: the ability to explore the area to find unknown targets and the ability to efficiently balance multiple (often conflicting) goals.
More specifically, a team of robots operate in an unknown environment searching for unknown hazardous targets in order to disarm them cooperatively. The locations of these unknown targets are detected gradually through searching by the robots. A target must be handled by a coalition of robots; therefore, the recruitment task starts in real time as the targets are found.

Once some of these targets are detected, a fixed number of the robots is needed to disarm the target through a predefined procedure of actions, while the remaining robots continue to explore the environment for searching other targets. Assuming that the robots make their decisions independently and that each of them, in a restricted region of the area, owns the same amount of information, each robot will decide in an autonomous  manner which of the two tasks to perform (explore or help the others in disarming process) and in what direction it may move. However, the search and coordination tasks are not entirely decoupled; it is possible for a robot to perform both simultaneously (for example when it moves towards the target, it also implicitly explores the area).

More specifically, the basic structure of our approach is divided into two steps:
\begin{enumerate}
\item  Detection, evaluation and selection of the cells, while moving during the exploration of the area.
\item Selection among different targets found so far in the area and navigation towards the chosen target.
\end{enumerate}
However the strategy is highly flexible allowing to reconsider, at each time step, the previous decisions taken at an earlier step to react to any new events that may occur.

\subsection{Environment modeling}

Let $A$ be the 2-D working field or grid, where $A \subset\mathbb{R}^2$. As a symbolic representation, the proposed method uses a grid map with $m$ and $n$ cells. Let us establish a Cartesian coordinate system taking the upper left corner of $A$ as the origin, each cell $c$ $\in$ $A$ of the area has its own definite coordinates $(x,y)$, with $x$ $\in$ $\{$$1, 2, \dots , m$$\}$ and $y$ $\in$ $\{$$1, 2, \dots, n$$\}$ elements.
The universal set $C$ contains all possible states of a cell on the grid map. The subsets $C_1, C_2, C_3,C_4$  $\subset$ $C$ (where $C_i \cap C_j = \emptyset$, $i\ne j$) represent possible states as follows:
\begin{itemize}
	\item {$C_1$}:$\{$explored by the robots$\}$,
	\item $C_2$:$\{$accessible and not explored by the robots$\}$,
	\item $C_3$:$\{$occupied by an obstacle$\}$,
	\item $C_4$:$\{$not explored and inaccessible after hazardous events (e.g., the mine's explosion) $\}$.
\end{itemize}

\begin{figure}
	\flushleft
	\includegraphics[scale=0.5]{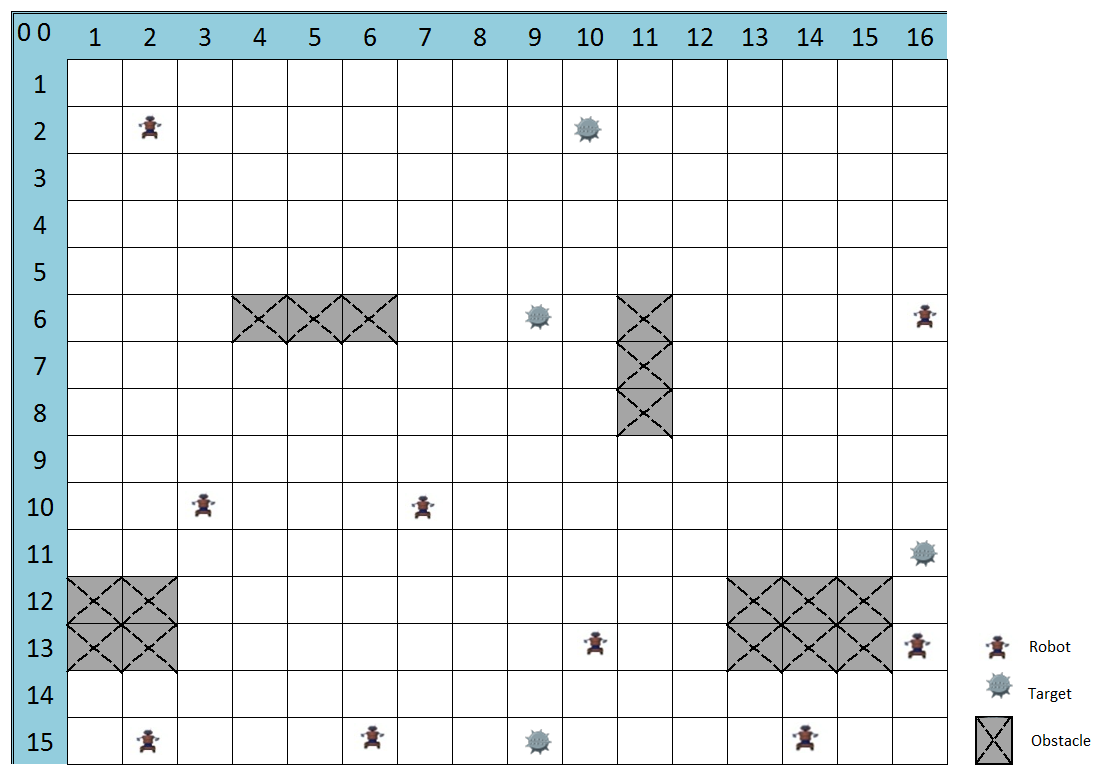}
	\caption{A representation of the simulation environment.}
	\label{fig:GridArea}
\end{figure}

Obstacle cells are inaccessible to the robots and impenetrable to the sensors. A cell occupied, at time step $t$, by any of the robots is considered as an obstacle, thus no other robot can take the place (see Fig.~\ref{fig:GridArea}).

The state $C_4$  is used in the dynamic scenario, as descried below, where the found targets such as mines could explode or dangerous chemicals may leak at any time, making the nearby cells inaccessible.

\begin{figure}
	\flushleft
	\subfloat[][\emph]
	{\includegraphics[width=.4\textwidth]{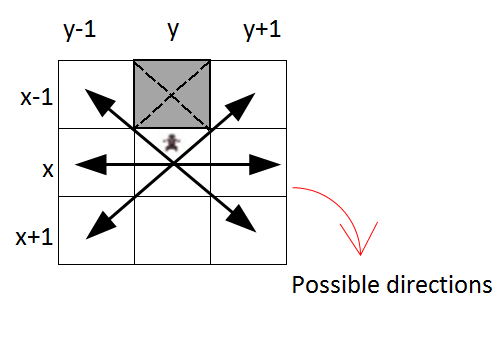}} \quad
	\subfloat[][\emph]
	{\includegraphics[width=.4\textwidth]{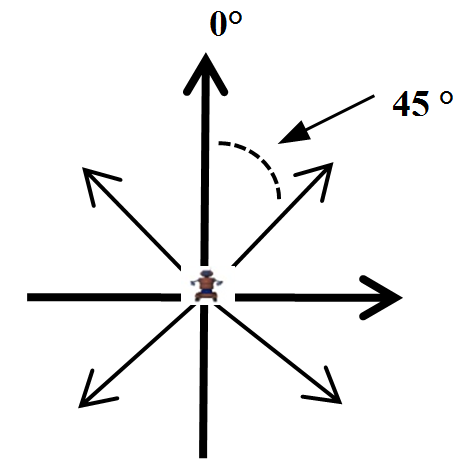}} \quad
	
	\caption{(a) Possible directions of a robot's move (b) Possible angles of
a robot's turn.}
	\label{fig:RobotDirections}
	
\end{figure}

While the robots explore the area, the cells transit to subset $C_1$.
Each cell $c = (x,y)$ has a maximum of eight adjacent neighbors $N(c)$, if all cells are accessible: $(x-1, y-1), (x-1,y), (x+1,y), (x,y+1), (x,y-1) (x+1, y+1), (x-1, y+1), and (x+1,y-1)$ as shown in Fig. \ref{fig:RobotDirections}.

\subsection{Assumptions}

A set R of homogeneous robots can be deployed in the area, where R = $\{$ $k$ $\mid$ $k$ $\in$ $\{$$1, 2, \dots ,$ $N^R$$\}$$\}$. At each step $t$, the current state of a robot $k$ can be represented by its coordinates $(x_k^t, y_k^t)$.
The robots are modeled as rational collaborative autonomous agents that move autonomously in the environment.  We assume that these robots are identical (executing the same algorithms) and follow simple local rules to communicate with neighbors and with the environment in order to provide a scalable strategy. However, for the sake of the simplicity, the robots
are equipped with advanced devices such as sensors, global positioning capabilities (for instance they are equipped with a Wi-fi module) camera, radar, and an onboard automatic target recognition system, with which the robots identify the targets such as mines and obstacles or other robots in proximity. Sensor's information is assumed to be perfect, and we assume that the robots have perfect knowledge of their locations expressed in terms of their coordinates.

\begin{figure}
	\flushleft
	\includegraphics[scale=0.4]{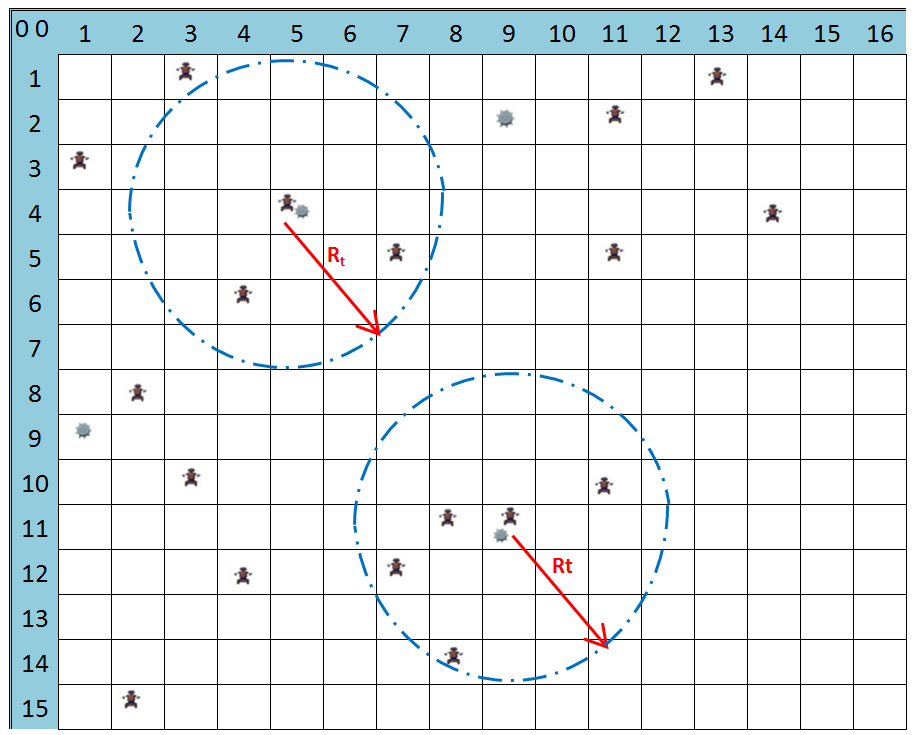}
	\caption{The robots in the cells with coordinates (4,5) and (11,9) have detected a target. They start a recruitment process by sending packets that will be received by the robots within their wireless range $R_t$. }
	\label{fig:WirelessRange}
\end{figure}

They are able to communicate with others using wireless communication and the communication range $R_t$ is limited, compared to the size of area, so two robots can exchange information only if they are close enough; i.e., the distance between them is smaller than $R_t$. We define a local neighborhood of robot $k$ at time $t$, denoted by $LN_k^t$ , as the set of  robots that are within the $R_t$ of the robot $k$ (Fig.~\ref{fig:WirelessRange}).

In addition, it is also assumed that the communication network is perfect (no packets loss, no transmission time or delay), so robots within the same wireless range have identical information at the same time. For the ease of presentation, the robots start the search simultaneously at the same time. These assumptions, that can easily be removed in the future, but for the moment this is to simplify the model since we are more interested in analyzing the trade-off in the proposed model.

The robots must explore the area for searching and dealing with a set T of $N^T$  targets such as mines disseminated in the area, i.e.,  T = $\{$ $z$ $\mid$ $z$ $\in$  $\{$1, 2, \dots , $N^T \} \}$. It is assumed that there is no prior knowledge about these targets such as the total number and locations. Thus the targets can be located in any position of the area with the same probability. Therefore the only way to ensure the detection of all targets (mines) is to do a complete search of the area.

Each target, is represented by its coordinates $(x_z, y_z)$. A target \textit{z} is detected by a robot $k$ when the target's coordinates coincides with the robot's  coordinates. Once a robot finds a target, it starts a recruitment process since a  target requires a fixed number of robots to be handled. We define $R_{\min}$ a non-negative integer represents the number of robots needed to treat safely a target.
 For this purpose, it exchanges wireless messages locally by sending out help requests through packets (that contains mainly the coordinates of the found target) to the robots within its wireless range $R_t$. We denote $RR_k$ as a set that keeps track of the help requests that the robot $k$ receives from the others, expressed in terms of found targets, thus $RR_k$ $\subset$ $T$.

\begin{figure}
	\flushleft
	\includegraphics[scale=0.4]{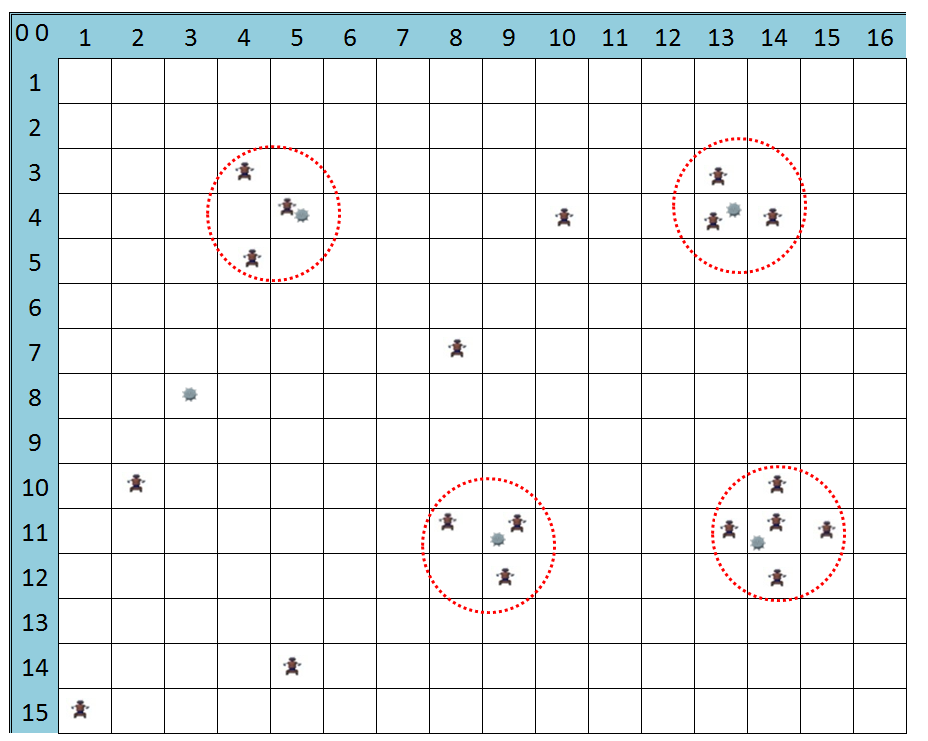}
	\caption{Local coalition of robots formed through the recruitment process}
	\label{fig:LocalCoalition}
\end{figure}

Figure \ref{fig:LocalCoalition} shows an example of local coalitions that are formed through the recruitment's processes. Since the robot's decisions can dynamically be changed in terms of robot's movements, new requests, failures, etc. the final configurations in the target's locations can change.

\subsubsection{The actions of robots}

The actions of robots belong to three main classes:
\begin{enumerate}[label=\Roman{*}., ref=(\Roman{*})]
	\item Sensing actions that  affect changes in a robot's knowledge of the environment.
	\item Moving actions in the cells which imply rotation to choose the right directions and  obstacles avoidance.
	\item Communication actions when targets are found.
\end{enumerate}

Each robot adapts its position in three different ways. The first is in the direction of minimum amount of pheromone (to indicate good feasible regions unexplored). The second is to move away from other robots or obstacles (to avoid collision). The third is in the direction of the detected targets (to perform an disarming task cooperatively). The first two are based on interactions accumulated over time between the robots and the environment. The third is a reactive behavior triggered by help requests from other robots. In the following, these behavior characteristics are described.

To behave as a collective robotic organism, the robots need to be able to achieve different behavioral states. They are able to reconfigure themselves so as to achieve a transition between the states. More specifically, at the beginning, when no target is detected,  each robot collects information from its immediate surrounding cells perceiving chemical substance (pheromone) by on-board sensors and uses this information to identify the direction where to move. Each robot calculates its best move as the next position locally according to an Ant Colony approach as explained below. The goal is that the robots should explore the undetected sub-area as much as possible in order to speed up the task. This state is named the  Forager State and it is the initial state for each robot.

Once a robot discovers a target by itself, it will switch to a Coordinator State.  Each coordinator robot is responsible for handling the disarmament process of the discovered target and for the recruitment of the others.  The recruiting process starts by broadcasting packets to the robots in its wireless range  (see Fig.~\ref{fig:WirelessRange}), and it ends when the predefined number of necessary robots ($R_{min}$) arrived to the target's location to form a coalition team. Then, the accumulated robots work together as a group, performing the disarmament task.
Essentially a coordinator robot performs the following steps:
\begin{enumerate}
	\item Check if there are a sufficient number of robots to form a coalition to handle a target.
	\item If there is no coalition that satisfies the constraint, then continue to send packets.
	\item Repeat step 2 until all necessary robots are arrived.
	\item Otherwise, stop the communication and start to disarm the target properly.
\end{enumerate}
Once the target is disarmed, the involved robots return to continue to explore the area.

\begin{figure}[t!]
	\subfloat[][\emph]
	{\includegraphics[width=.45\textwidth]{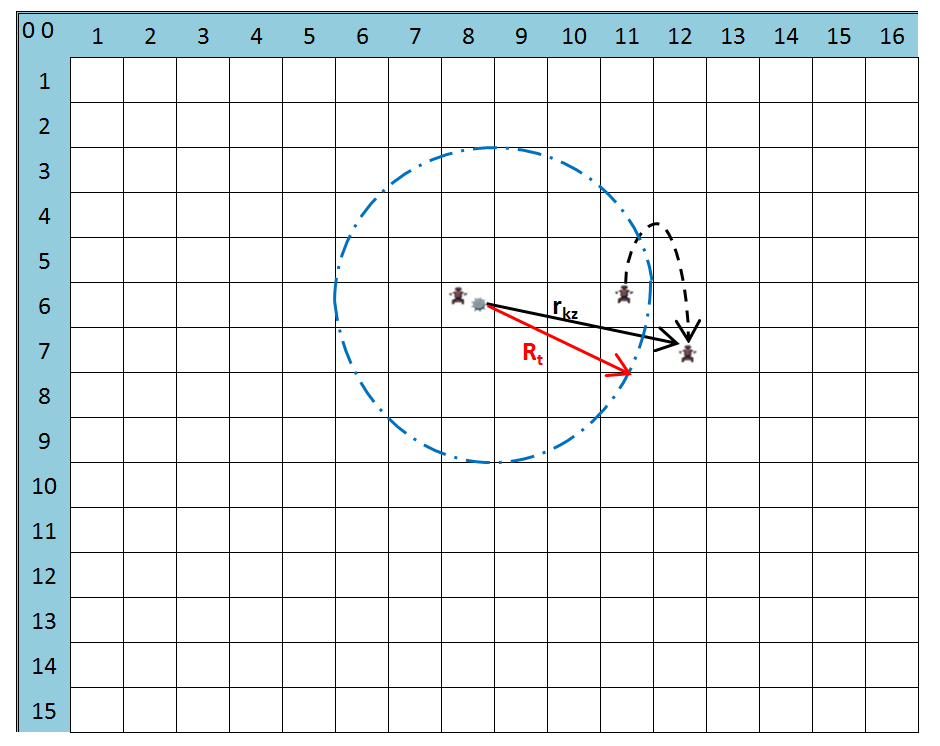}} \quad
	\subfloat[][\emph]
	{\includegraphics[width=.45\textwidth]{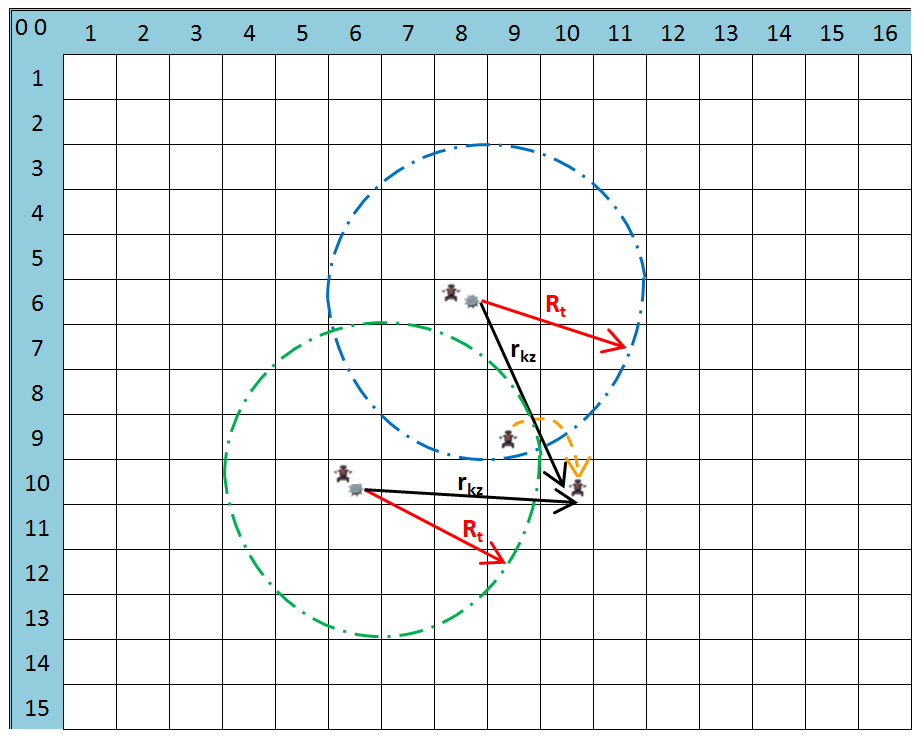}} \quad
	\caption{(a) The robot in the cell (6,11) that is recruited by the robot in cell (6,8) moves into a cell that is too far from the target, thus it changes its state by becoming an explorer. (b) The robot in cell (9,9) that is recruited by two robots in both cells (6,8) and (10,6), respectively. After, it moves into the cell that is too far from both targets, thus it changes its state by becoming a forager.}
	\label{fig:ChangeState}
\end{figure}

When a robot $k$ receives one or more request packets by coordinator robots, it will make the decision whether to continue to explore or to help in a disarming process. If it decides to help a coordinator robot, it forms a list  $RR_k$ of the received packets in terms of targets; otherwise, the task is rejected and the robot continues to explore individually. If there is only one request, it has to decide whether it should move to remove the target or continue to explore the area. If more requests have been received, it not only needs to decide whether it should remove, but also it has to decide towards which target it could move.  This happens because a robot has to balance the two tasks according to a weighted objective function express below. This state is named the Recruited State.

A key aspect of this state is that the robots react to events that occur. Unlike common approaches, they could change the decisions taken previously during the iterations. For example, for a certain type of mission, it is possible to meet a target or receive different requests, while reaching another target in response to a recruitment process, thus reconsidering the choice of the target to be handled. Moreover, the decision can be to restart to explore the area since the movements are too far from the target's location see Fig.~\ref{fig:ChangeState}.

When a recruited robot, once it reaches the target's location, it will wait until the other needed robots to arrive and thus enter into the waiting mode. This state is called the Waiting State.

Finally, once the number of needed robots reach the target's location, the group as a whole is involved in the disarming process and they will perform, for a fixed amount of time, some actions to deal with the targets properly. This state is the Execution State.

\begin{figure}
	\flushleft
	\includegraphics[scale=0.4]{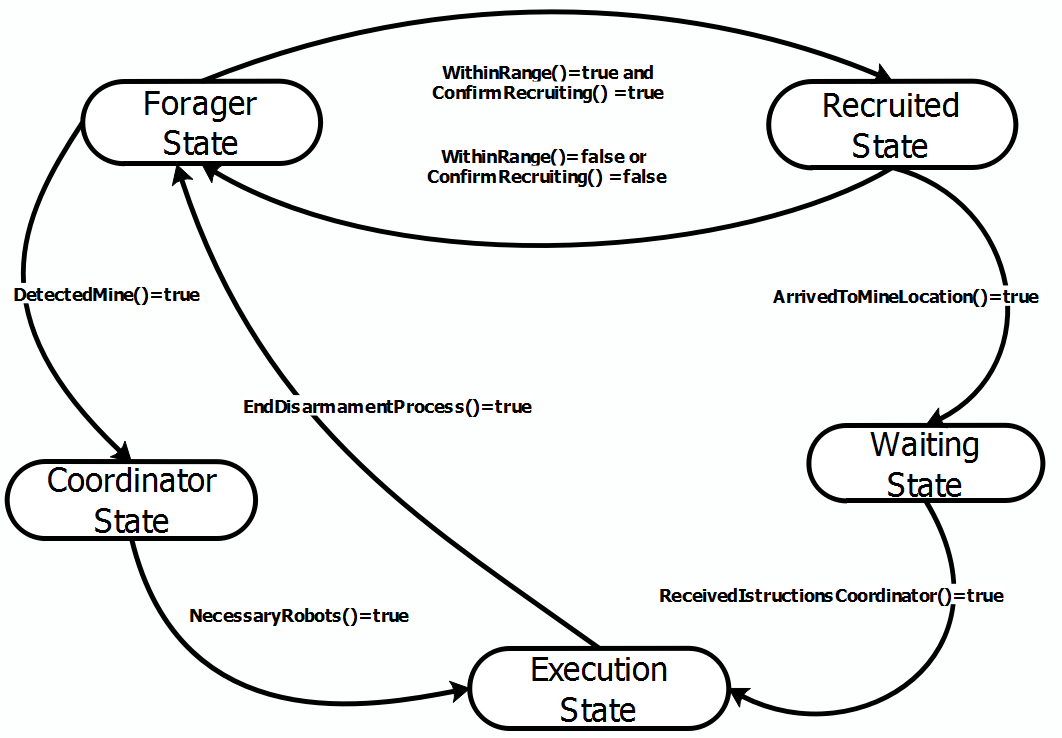}
	\caption{State transition logic for robots at each time step.
		WithinRange() is a function returning True when the robot is within
		the communication range of a coordinator robot. Confirm Recruiting() is a function returning True if the robot decides to get involved in the disarmament process of a target. ArrivedToMineLocation() and ReceivedIstructionsCoordinator() are two functions returning True if the robot is arrived to a target's location and received the command to start the disarmament process by the coordinator, respectively. NecessaryRobot() returns True if all needed robots ($R_{min}$) have arrived at the target's location.}
	\label{fig:StateTransition}
\end{figure}

To summarize the above actions and states, Fig.~\ref{fig:StateTransition} shows the state transition logic of robots at each time step.

\section{Multi-Objective Optimization Formulation}

Multiple conflicting objectives may arise naturally in most real-world robotic optimization problems. Several principles and strategies have been developed and proposed for over the last decades in order to solve such problems. In multi-objective optimization, as its name implies, there are multiple objective functions with a possibility of conflicting with each other. The aim is to find a set of vectors of decision variables that can satisfy constraints and optimize (minimizes or maximizes) these functions. Such solution vectors are not a unique vector, there are many such solutions vectors forming a so-called Pareto front. Each point or non-dominated solution on the Pareto front provides a preference and choice between different objectives. When the Pareto front becomes convex, weighted sum methods can aggregate different objectives into a single objective.

In general, a multi-objective optimization problem can be written mathematically as
\begin{multline}
\label{ObjectiveFunction}
\textrm{To find vectors}
\hspace{0.3cm} X = (x_1, x_2,\dots, x_L)^T  \in \Omega \\
\textrm{which optimize}
\hspace{0.3cm}	f(x) = (f_1(x), f_2(x), \dots, f_p(x)) \\	
\textrm{subject to}
\quad g_j(x)\le 0,  \; j=1, \dots, s, \quad \\
h_r(x)=0, \; r=1,...,d,  \qquad \qquad \qquad \qquad \qquad \qquad  \; 	
\end{multline}
where $f_1(x), f_2(x), \dots, f_p(x)$ denote the objective functions to be optimized simultaneously, $X$ is the vector of the decision variables in the search/decision space. $\Omega$ is the set of feasible solutions and $g_j(x)$ denotes the inequality constraints,
while $h_r(x)$ are equality constraints. All these functions can be
linear or nonlinear \cite{39}.

As it is very difficult to effective handle with all the conflicting objective functions, several methods have been developed for this purpose. One of these methods is that the multi-objective problem is transformed into a single-objective problem by a weighted sum.
In this paper, in order to solve our bi-objective problem, the weighted sum method is used
to deal with conflicting goals and the solutions can be obtained as a trade-off of the specific problem. The total cost of the fitness function is obtained by a linear combination of the weighted sum of two objectives in which each objective function based on its importance or preference \cite{40}.

The problem is transformed into a single-objective optimization problem by using scalar weighting factors associated with each objective function as follows:
\begin{equation}
\label{eq: FunctionWeighted}
F_{\textrm{weighted\; sum}} = \sum_{i=1}^{p} w_if_i(x)
\end{equation}
where $w_i$
\begin{equation}
\label{eq: WeightFactorCondition}
w_1+w_2+, \dots w_p = 1, \quad w_i \ge 0.
\end{equation}

The weighted sum method changes weights systematically, and each different single objective optimization determines a different optimal solution.  This approach gives an idea about the shape of the Pareto front and allows information to be obtained about the trade-off among the various objectives to accumulate gradually \cite{41}.

\subsection{Formulation as a Bi-objective Problem}

In order to define the problem as a bi-objective optimization model, there are different
decisions to be made by a robot. Given the position, expressed by the coordinates, where each robot (e.g., robot $k$) is located at each time step, and given a found target $z$ $\in$ $RR_k$, then the robot $k$ has to decide if it will get involved in the recruiting process or will continue to explore the area.

\subsubsection{First Objective}

Firstly, the first decision can mathematically be represented as the following decision variable:	
\begin{equation}
\label{eq: ExplorationVariable}
v_{xy}^k=
\begin{cases}
1  & \text { if the robot $k$ visits the cell ($x,y$), } \\
0  & \text { otherwise.}
\end{cases}
\end{equation}

We assume that the time to visit a cell is denoted by $T_e$ and it is supposed to be the same for all robots. Then the goal of an exploration task is to cover the whole area in the minimum amount of time, and thus the first objective becomes
\begin{equation}
\textrm{minimize } \sum_{k=1}^{N^R}\sum_{x=1}^{m} \sum_{y=1}^{n}  T_{e}\ v_{xy}^k.
\end{equation}

\subsubsection{Second Objective}

Similarly, the following decision variable allows us to model if a robot $k$ is involved in the disarmament process of a found target $z$:
\begin{equation}
\label{eq: TargetVariable}
u_{z}^k=
\begin{cases}
1  & \text { if robot $k$ is involved with target $z$, } \\
0  & \text { otherwise. }
\end{cases}
\end{equation}

When a robot has eventually detected a target, it should act as an attractor, trying to recruit the required number of robots so as to disarm the discovered target safely and properly.
As mentioned above, it is necessary to form a coalition to handle a target. We define the number of robots that can deal with a target as $R_{min}$. Therefore, it sends help requests using wireless communication via packets that are received by the other robots in the wireless range. It should be noted that the received packets are not forwarded, since we are focused on one hop communication. Then, the decision is taken by the recruited robots. The quality of the recruitment process is measured in terms of time.

Let $T_{Start,z}^k$ be the time step at which the robot $k$ receive a help request for disarming the target $z$ and $T_{End,z}^k$ the time step at which the robot $k$ has reached the target $z$, then ($T_{End,z}^k$ - $T_{Start,z}^k$) is the time taken for coordination, namely, the coordination time. Thus, the objective is the minimization of the coordination time for each found target, in order to speed up the disarming process and continue the mission effectively.
Therefore, the second objective is
\begin{equation}
\textrm{minimize } \sum_{k=1}^{N^R}\sum_{z=1}^{N^T}  (T_{End,z}^k-T_{Start,z}^k)\ u_{z}^k.
\end{equation}

\subsubsection{The Bi-Objective Optimization Problem}
The considered objective function is thus related to the minimization of the time needed to perform the overall mission. The problem of selecting the best solutions for the problem (in the Pareto sense), accounting both the exploration time and the coordination time, can be mathematically stated as follows:
\begin{multline}
\label{ObjectiveFunction}
\textrm{minimize }
\sum_{k=1}^{N^R}\sum_{x=1}^{m} \sum_{y=1}^{n}  T_{e}\ v_{xy}^k,  \quad \textrm{and }
\textrm{minimize } \sum_{k=1}^{N^R}\sum_{z=1}^{N^T}  (T_{End,z}^k-T_{Start,z}^k)\ u_{z}^k, 	
\end{multline}
subject to
\begin{equation}
\label{eq: ConstraintCell}
\sum_{k=1}^{N^R} v_{xy}^k \geq 1, \quad \forall\ (x,y) \in A,
\end{equation}
\begin{equation}
\label{eq: ConstraintTarget}
\sum_{k=1}^{N^R} u_{z}^k =  R_{min},  \quad \forall \ z \in T,
\end{equation}
\begin{equation}
\label{eq: DomainVariableCell}
v_{xy}^k \in \ \{ 0,1 \}, \quad  \forall \ (x,y) \in A, \  k \in R,
\end{equation}
\begin{equation}
\label{eq: DomainVariableTarget}
u_{z}^k \in \ \{ 0,1 \}, \quad \forall \ z \in A,\  k \in R,
\end{equation}
\begin{equation}
\label{eq: DomainTimeVariable}
T_e, \ T_{End,z}^k,\  T_{Start,z}^k \in R, \quad \forall \ z \in A,\  k \in R.
\end{equation}

The objective functions in (\ref{ObjectiveFunction}) to be minimized represent the total time consumed by the swarm of robots. They depend on both the time for the exploration of the area and the time for coordinating the robots involved in the disarming process of the targets. Constraint (\ref{eq: ConstraintCell}) ensures that each cell is visited at least once. Constraint (\ref{eq: ConstraintTarget}) defines that each target $z$ must be disarmed safely by $R_{min}$ robots.
The constraints (\ref{eq: DomainVariableCell})-(\ref{eq: DomainTimeVariable}) specify the domain of the decision variables.

The problem was formulated as a bi-criteria model which turns out to be very challenging to solve. Indeed, the number of efficient solutions may be exponential in terms of the problem size, thus prohibiting any efficient method to determine all efficient solutions. For these reasons, following the popular approaches used to deal with multi-objective optimization problems, the model has been transformed into a single objective optimization problem using arbitrary importance factors for each criteria ( i.e. $w_1$ and $w_2$) and combining the objectives as a single function to be minimized. The resulting single objective problem with non-negative weights can be represented as follows:
\begin{multline}
\label{ObjectiveFunction1}
\textrm{minimize }
\sum_{k=1}^{N^R}\sum_{x=1}^{m} \sum_{y=1}^{n} w_1( T_{e}\ v_{xy}^k)
+ \sum_{k=1}^{N^R}\sum_{z=1}^{N^T} w_2 [(T_{End,z}^k-T_{Start,z}^k)]\ u_{z}^k, 	
\end{multline}
subject to constraints (\ref{eq: ConstraintCell})-(\ref{eq: DomainTimeVariable}).

Parameters $w_1$ and $w_2$ are chosen such that the condition $w_1+w_2=1$ is satisfied.
 In this case, the combined function is Pareto optimal \cite{41}. The user can choose appropriate values for the parameters $w_1$ and $w_2$, depending on the preference or priority of the objectives. Indeed, by minimizing the weighted sum objective with various settings, it is possible to determine various points in the Pareto set. This approach can approximate and describe the shape of the Pareto front effectively, allowing the accumulation of information to be obtained about the trade-off among various objectives.

The proposed single objective optimization model can be solved and be relevant to many applications for robot exploration and coordination. For applications in which more relevance is given to the exploration task, more importance could be given to exploration time
(thus higher value of $w_1$), whereas for applications where it is more important to reduce the disarming time, more importance could be given to $w_2$.

 By minimizing the overall fitness function in regard to the assigned weights of each criterion, a suitable decision mechanism that may balance the two objectives can be obtained. The weights have been tuned through a set of simulations in order to try to find the best values.

\subsection{Energy Model}
For each activity executed by a robot $k$, a certain amount of energy is consumed. In our study, the energy model reflects mostly two activities: energy for communication and energy for mobility. The mobility energy depends on several factors. For simplicity, the mobility cost for a robot $k$ in our model can be calculated by considering the distance traversed in terms of the number of cells and it is expressed as follows:
\begin{equation}
\label{eq: ExplorationEnergy}
E_{mob}^k=\sum_{x=1}^{m} \sum_{y=1}^{n} \ C_{mob}\ v_{xy}^k,
\end{equation}
where $\sum_{x=1}^{m} \sum_{y=1}^{n}$  $v_{xy}^k$  is the total number of visited cells for each robot $k$ while moving in the exploration phase and recruiting phase; $C_{mob}$ is the energy cost given to move from one cell to another and takes into account both the costs for moving and turning \cite{42}.

When a target is detected, the energy consumed is instead related to the communication and to the cost for performing the planned task on the target. Since we use a wireless communication system in this phase,  the energy consumed depends on the transmission and reception of the packets to communicate the position of the targets. In this case, we assume that the energy consumed by robot $k$ to transmit $E_{tx}^k$ and receive $E_{rx}^k$ a packet \cite{43} is related to the maximum transmission range $R_t$ and to the packet size $(l)$ as follows:
\begin{equation}
\label{eq: TrasmissionEnergy}
E_{tx}^k=l\ (R_t^\psi\ e_{tx}+ e_{cct}),\
\end{equation}
where $e_{tx}$ is the energy required by the power amplifier of transceiver to transmit one bit data over the distance of one meter, and $e_{cct}$ is the energy consumed in the electronic circuits of the transceiver to transmit or receive one bit. Here, $\psi$ is called the path loss exponent of the transmission medium where $\psi$ $\in$ $[2, 6]$.

On the other hand, the energy consumption for receiving a packet is independent of the distance between communication nodes and it is defined as:
\begin{equation}
\label{eq: ReceivingEnergy}
E_{rx}^k=l\ e_{rc},\
\end{equation}
The energy consumed to deal with a target is:
\begin{equation}
\label{eq: PerformingEnergy}
E_{d}^k=C_d,\
\end{equation}
where $C_d$  is the cost given to the working task for handling a target properly, and it is the same for each robot and it is related, for simplicity, to the mechanical movements.
Essentially, we model the energy consumed for the coordination task by the robot $k$ that is involved in the targets issue as:
\begin{equation}
\label{eq: CoordinationEnergy}
E_{coord}^k=\sum_{z=1}^{N^T} (E_{tx}^k + E_{rx}^k+ E_{d}^k) \ u_{z}^k.
\end{equation}

Based on the previous considerations and models, we now introduce a performance index, called Total-Energy-Swarm-Consumption (TESC), as:
\begin{equation}
\label{eq: aesc}
TESC=\sum_{k=1}^{N^R}\ E_{mob}^k + \sum_{k=1}^{N^R} E_{coord}^k.
\end{equation}
That is, the total energy consumed by the swarm of robots and it is the sum of two contributions: energy consumption for moving into the area and energy consumption for the wireless communication when they are involved in the performing of the targets.

\subsection{Robot in dynamic scenarios}

The above considerations provide a unified approach to consider both the complete discovery of the area and the measure of the time needed to accomplish both exploration and disarming of the targets. Thus, it is a useful metric, but it requires that the task is completely finished, and cannot be used to evaluate partial execution of the tasks. In many cases a complete exploration of the environment may not be feasible in practice, due to the time or resource constraints in large and hazardous environments.

In this section, we consider  a dynamic environment in the sense that the targets can explode at any time and in an unpredictable manner, mimicking the destruction of some robots and the damage of the nearby zones. Moreover, we regard the robots with a limited quantity of energy without the possibility of recharge or replacement. In such scenarios, the team works under more demanding time constraints.

In order to use a performance metric that is applicable to the robotic system in a dynamic scenario, we have considered several functions; one for each feature that must be discovered and measured from the environments. More specifically, the performance metrics are given by each function measuring the percentage/ratio of information related to the two tasks. In the case of exploration task, it is the percentage of the environment explored  not covered by impassable obstacles, while in the case of the disarmament task, it is the percentage of targets successfully identified and disarmed.

The following equations summarize the region of an emergency scene as follows:
\begin{equation}
A_E= \sum_{x=1}^m\sum_{y=1}^n c_{(xy)}     \\  c_{xy} \in 	C_1,
\end{equation}

\begin{equation}
A_{UN}= \sum_{x=1}^m\sum_{y=1}^n c_{(xy)}     \\  c_{xy} \in 	C_2, C_4.
\end{equation}

Concerning the above regions of interest, we define the following terms:
\begin{equation}
F_1=\frac{ A_E}{\sum_{x=1}^m\sum_{y=1}^n  \\ c_{(xy)}},  \quad   c_{xy} \in 	C_1,C_2, C_4
\end{equation}
where $F_1$ is a regularized term that indicates the percentage of explored cells in the emergency scene. Thus, $F_1$ will be equal to one only in the case all cells of the area have been explored, except for the cells with obstacles $(c \in C_3)$.

Now we define the number $F_2$ of handled targets as follows:
$$ F_2= \sum_{z=1}^{N^T} f(z) $$
\begin{equation}
\label{eq: numberofdisarmedmine}
=
\begin{cases}
1  & \text { if target z is disarmed properly, } \\
0  & \text { otherwise.}
\end{cases}
\end{equation}

In this case, the objectives essentially become the maximization of the percentage of explored area and the number of disarmed targets. In this case,  the robots have a limited amount of energy and at each time step, a fixed quantity of energy is consumed (see Section 4.2) depending on what action the robot may perform and if the mines can explode. The mission can terminate for multiple reasons, including the case that all robots have used up the energy, or are damaged due to explosion.

\section{Proposed Bio-Inspired Approaches}

Our proposed approach to solve the problem is a hybrid strategy which combines both indirect and direct communication mechanisms. We study how robots can accomplish the mission in a distributed and self-organized way through a stigmergic process in the exploration task, and simple information locally sent by the robots in the recruitment task.
Our system has unique features such as the minimal information exchange, and local interactions between simple homogeneous robots, achieving complex collective behaviour. Such solutions are in line with the general approaches used in swarm robotics, and support the desired system properties of robustness, adaptivity and scalability.

\subsection{Repulse Pheromone-Based Strategy for Exploration}

The mobile, autonomous robots, performing the exploration search task, must be able to decide the sequence of movements needed to explore the whole environment.
In this work, we address the exploration problem in the context of search and rescue operations. Exploration strategies that drive the robots around an unknown environment on the basis of the available knowledge are fundamental for an effective search. The mainstream approaches for developing exploration strategies are mostly based on the idea of incrementally exploring environment by evaluating a number of candidate observation locations, in our cases neighbor cells, according to a criterion and by selecting, at each step the next best location. However, we do not address the problem to build a map of the environment, since we are more interested in locating the largest number of targets in the minimum amount of time. Differently from map building, search and rescue settings are strongly constrained by both time and battery limitations and generally require the amount of explored area over the map quality. Since the robots should be required to be capable of various functionalities other than area exploration, it is desirable that both the integration to a swarm and the ability to explore are seamless and these actions should not consume a large amount of the robot’s resources.

To be effective, a search strategy must attract robots towards unobserved areas so as to avoid the undesirable scenario where some areas are frequently revisited while others remain unexplored. Therefore, some swarm control is needed. The swarm control algorithm used here is a pheromone-inspired mechanism in which the environment is assumed to handle the storing and diffusion of chemical substance; thus, the robot controllers do not store any chemical information, except the sampled concentrations within the immediate vicinity of the robot. Utilizing a stigmergic communication would be an efficient method of achieving such emergent behaviour with low overhead.

Essentially, when the robots are exploring the area, they lay pheromone on the traversed cells and each robot uses the distribution of pheromone in its immediate vicinity to decide where to move. Like in nature, the pheromone trails change in both space and time.
The pheromone deposited by a robot on a cell  diffuses outwards cell-by-cell until a certain distance $R_s$ such that $R_s \subset A \subset\mathbb{R}^2$ and the amount of the pheromone decreases as the distance from the robot increases (see  \figurename~\ref{fig:DiffusionPheromone}).

Mathematically, the pheromone diffusion is defined as follows: consider that robot $k$ at iteration $t$ is located in a cell  of  coordinates ($x_k^t$, $y_k^t$) $\in$ $A$, then the amount of pheromone that the robot deposits at the cell $c$ of coordinates $(x,y)$ is given by:

\begin{figure}
	\center
	\includegraphics[scale=0.4]{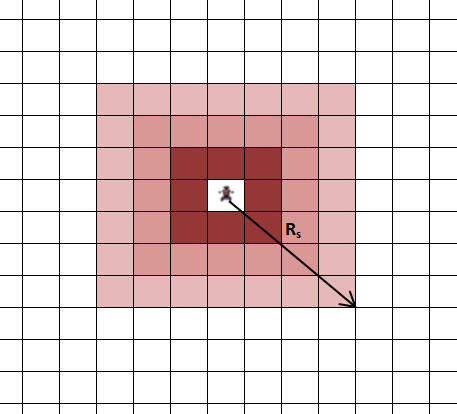}
	\caption{Example of pheromone diffusion. When a robot moves into new cell, it spreads the pheromone within a certain distance $R_s$. The intensity of pheromone decays according to the distance from the cell. }
	\label{fig:DiffusionPheromone}
\end{figure}

\begin{equation}
\label{eq: PheromoneCell}
\Delta\tau_{c}^{k,t} =
\begin{cases}
\Delta\tau_0\ e^\frac{- r_{kc} }{a_1} - \frac{\varepsilon}{a_2}  & \text {if  $ r_{kc}$  $\le$ $R_{s}$,   }  \\
0 & \textrm{otherwise,}
\end{cases}
\end{equation}
where $r_{kc}$ is the distance between the robot $k$ and the cell $c$ and it is defined as:
\begin{equation}
r_{kc} = \sqrt{(x_k^t-x)^2 +(y_k^t-y)^2}.
\end{equation}
This means that pheromone spreads up to a certain distance, as in the real world, after which it is no perceivable by other robots. In addition, $\Delta$$\tau_{o}$ is the quantity of pheromone sprayed in the cell where the robot $k$  is placed and it is the maximum amount of pheromone,  $\varepsilon$ is an heuristic value (noise) and $\varepsilon$ $\in$ $(0,1)$. Furthermore, $a_1$ and $a_2$ are two constants to reduce or increase the effect of the noise and pheromone. It should be noted that multiple robots can deposit pheromone in the environment at same time, then the total amount of pheromone that can be sensed in a cell $c$ depends on the contribution of many robots.

Furthermore, the deposited pheromone concentration is not fixed and evaporates with the time. The rate of evaporation of pheromone is given by $\rho$ ($0 \le \rho \le 1$
), and the total amount of pheromone evaporated in the cell $c$ at step $t$  is given by the following function:
\begin{equation}
\label{eq: pheromoneEvaporated}
\xi_{c}^t = \rho\ \tau_{c}^t,
\end{equation}
where $\tau_{c}^t$ is the total amount of the pheromone on the cell $c$ at iteration $t$.

Considering the evaporation of the pheromone and the diffusion according to the distance, the total amount of pheromone in the  cell $c$ at iteration $t$ is given by
\begin{equation}
\label{eq: totalPheromone}
\tau_{c}^t = \tau_{c}^{(t-1)} - \xi_{c}^{(t-1)} + \sum_{k=1}^{N^R} \Delta\tau_{c}^{k,t}.
\end{equation}

\subsection{Cells Selection}

At each time step, the algorithm selects the most appropriate cell for each robot, among a set of neighbor cells without the knowledge of the entire area. This happens because the robots have not got global information about the environment.
The aim is to avoid any overlapping and redundancy efforts, therefore, the robots must be highly dispersed in the area in order to  complete the mission as quickly as possible, avoiding at the same time any wastage of the robot's resources such as energy.

Each robot $k$, at each time step $t$, is placed on a particular cell $c_k^t$ that is surrounded by a set of accessible neighbor cells $N(c_k^t)$. Essentially, each robot perceives the pheromone deposited into the nearby cells, and then it chooses which cell to move to at the next step. The probability at each step $t$ for a robot $k$ of moving from cell $c_k^t$ to cell $c$  $\in$ $N(c_k^t)$ can be calculated by
\begin{equation}
\label{eq: Probabilitychhosecell}
p(c|c_k^t) = \frac{(\tau_{c} ^t)^\varphi \  (\eta_{c} ^t )^\lambda       } {\sum_{b \in N(c_k^t) }   (\tau_{b} ^t)^\varphi \  (\eta_{b} ^t )^\lambda}, \quad \forall\ c \in N(c_k^t),
\end{equation}
where $(\tau_{c} ^t)^\varphi$ is  the quantity of pheromone in the cell $c$ at iteration $t$, and  $(\eta_{c} ^t )^\lambda $ is the heuristic variable to avoid the robots being trapped in a local minimum. In addition, $\varphi$ and $\lambda$ are two constant parameters which balance the weight to be given to pheromone values and heuristic values, respectively. The robot $k$ moves into the cell that satisfies the following condition:
\begin{equation}
\label{eq: SelectedCell}
c = \arg \min [p(c|c_k^t)].
\end{equation}

In this way, the robots will prefer less frequently visited regions and it is more likely that it will direct towards an unexplored region.  The exploration strategy is detailed in Algorithm~\ref{ExplorationAlgorithm} and it was previously validated \cite{Palmieri2017}.
At the first iteration of Algorithm~\ref{ExplorationAlgorithm}, all cells are initialized with the  same value of the pheromone trail,  set to be zero that represents that the cells have not yet been visited by any of the robots, so that the initial probabilities that a cell  would be chosen is almost random.  Then the robots move from a cell to another  based on the cell transition rule in Eq.(\ref{eq: Probabilitychhosecell}).
Unvisited cells become more attractive to the robots in the subsequent iterations. Using this approach, the robots explore the area by following the flow of the minimum pheromone. Then the pheromone trails on the visited cells by ants are updated as in
Eq.~(\ref{eq: totalPheromone}).

\begin{figure}
	\flushleft
	\includegraphics[scale=0.5]{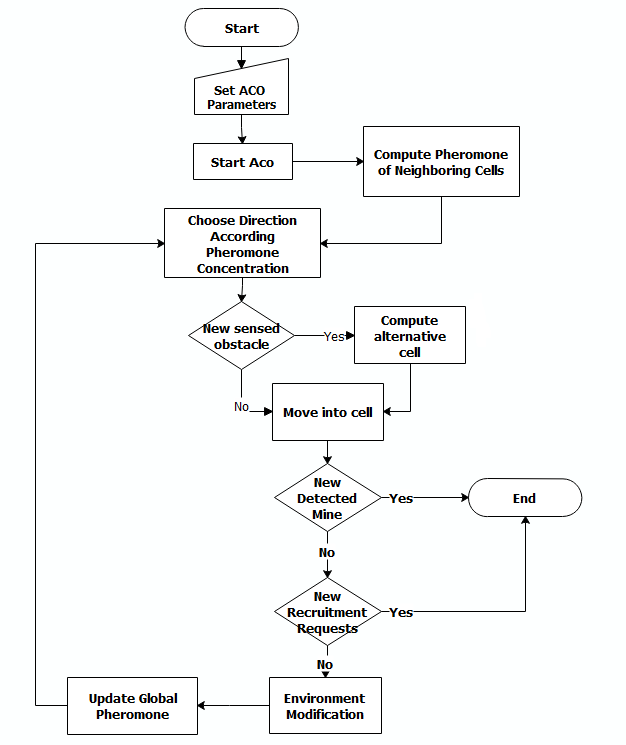}
	\caption{The flow chart of Exploration task for a robot}
	\label{fig:AcoFlowChart}
\end{figure}

Algorithm \ref{ExplorationAlgorithm} stops executing for a robot when it becomes a coordinator or it is recruited or if the mission is completed (that is all cells have been visited at least once and all targets are found and disarmed), therefore the frequency of its execution depends on the state of the robots during the mission.
Fig.~\ref{fig:AcoFlowChart} illustrates a simplified flowchart of the ACO-based strategy applied by each robot agent in exploration states.

\subsection{Recruitment task}

This task aims to design a low-cost coordination mechanism that is able to form groups of robots at given sites where the targets are found.  Once a robot detects a target (mine), since it does not have sufficient resource capabilities to handle it, it acts as a strong attractor to the other robots within the wireless range to form a coalition that cooperatively works for the disarmament process of the target.  The detection of a target may happen at any time during the exploration of the area, so the recruitment process can take place in different regions of the area.

\begin{algorithm}[t!]
	\caption{Exploration algorithm inspired by ant colony optimization. }
	\label{ExplorationAlgorithm}
	\LinesNotNumbered
	\SetAlgoLongEnd
	\SetKwData{Left}{left}\SetKwData{This}{this}\SetKwData{Up}{up}
	\SetKwFunction{Union}{Union}\SetKwFunction{FindCompress}{FindCompress}
	\Begin {
		\SetKwInOut{Stepi}{Step 1}
		\SetKwInOut{Stepii}{Step 2}
		\SetKwInOut{Stepiii}{Step 3}
		\SetKwInOut{Stepiiii}{Step 4}
		\Stepi{\textbf{Initialization.} \\
			Set $t$: $\{$$t$ is the time step$\}$. \\
			Define $\varphi$, $\lambda$, $a_1$, $a_2$, $\epsilon$, $\Delta\tau_0$, \\$\rho$, $R_s$ }
		\Stepii{\textbf{Generation coordination system.} \\For the whole swarm, set the \\initial locations in terms of\\ coordinates in $x$ and $y$ directions.}
		\Stepiii{\textbf{Procedure}}
		\While{the stop criteria are not satisfied}{
			\ForEach{robot $k$ in Forager State }{\label{forins}
				evaluate the current position $c_k^t$\;
				evaluate neighboorhood $N(c_k^t)$\;
				compute $c$ according Eq. (\ref{eq: SelectedCell})\;
				\eIf{($c$.hasObstacle() or $c$.isOccupated() or c.isInaccessible())}
				{choose a random cell $c^*$ $\in$ $N(c_k^t)$\;
					move robot $k$ towards $c^*$;}
				{move robot $k$ towards $c$;\\
					deposit pheromone according to Eq. \ref{eq: PheromoneCell};}
			}
			\ForEach{cell $c$ $\in$ $A$}{\label{forins}
				update pheromone according Eq.(\ref{eq: totalPheromone});
			}
			update $t$;
		}
	}
\end{algorithm}

For this purpose, wireless communication is used to share the information about the found targets, since direct communication may be beneficial when a fast reaction is expected and countermeasures must be taken. In this case, each robot is assumed to have transmitters and receivers, using which it can send packets to other robots within its wireless range $R_t$ and there is no propagation of the packets (one hop communication) as shown in Fig.~\ref{fig:WirelessRange}. The packets contain mostly the coordinates of the detected targets. Therefore, the volume of information that is communicated among the robots is small, but it implies the robots still lack global knowledge of the environment.

The most common approach is in a greedy fashion in which the target is instantaneously assigned to the robots without taking into account future events.  Here, we propose a flexible strategy in which the robots not only balance the two tasks such as exploration and manipulation of the targets (mine), but can react to future new events changing, eventually, the taken decisions. However, each robot must take individual decisions that could lead to retract itself from help requests. For example, for such kind of mission, it is possible to detect a target, while reaching another, or to receive another request, and thus may change decisions to move in a more convenient way from the robot's point of view. So at each time step, the robots will make the best selfish decision based on their positions, in response to the received help requests, trying at the same time to balance the two tasks.

It is worth mentioning that the decisions to be made by the robots are independent, and the other robots and the coordinators do not know the taken decisions; therefore, the coordinators robots will continue to send packets until the needed robots have actually arrived.

\subsubsection{Firefly-based Decision Mechanism for the Cooperation Task}

Firefly Algorithm (FA) is a nature-inspired stochastic global optimization method that was developed by Yang \cite{5}. It tries to mimic the flashing behaviour of a swarm of fireflies. A firefly in the search space communicates with the neighboring fireflies through its brightness which influences the selection.

Fireflies swarm in nature exhibit social behaviour that use collective intelligence to perform their essential activities like species recognition, foraging, defensive mechanism and mating. A firefly has a special mode of communication with its light intensity that signals to the swarm about its information concerning its species, location, attractiveness and so on. The two important properties of the firefly's flashing light are defined as follows:
\begin{itemize}
	\item brightness of the firefly is proportional to its attractiveness, and
	\item brightness and attractiveness of pair of fireflies is inversely proportional to the distance between two.
	\end{itemize}
These properties are responsible for the visibility of fireflies which pave way to communicate with each other.

In the proposed approach concerning our robot problem, each coordinator robot that has found a targets starts to behave like a firefly. At each time step, the probability of selection of a firefly is higher when the intensity value of the flashes is high. The intensity value depends on the distance. So the recruited robot moves towards the most attractiveness firefly/robot in the range. Attractiveness decreases as the distance increases.
The distance  $r_{ij}$ between any two fireflies $i$ and $j$, at positions $x_i$ and $x_j$, respectively, can be defined as the Euclidean distance as follows:

\begin{equation}
\label{eq: EuclideamDistance}
r_{ij} = ||x_i-x_j|| = \sqrt{ \sum_{d=1}^{D} (x_{i,d} - x_{j,d} )^2},
\end{equation}
where $x_{i,d}$ is the $d$th component of the spatial coordinate $x_i$ of the $i$th firefly and $D$ is the number of dimensions.
In 2-D case,  $r(i,j): \mathbb{R}^2$ $\rightarrow$ $\mathbb{R}$
\begin{equation}
r_{ij}= \sqrt{(x_i-x_j)^2+(y_i-y_j)^2}.
\end{equation}

In the firefly algorithm, as the attractiveness function of a firefly $j$ varies with distance, one should select any monotonically decreasing function of the distance to the chosen firefly. For example, we can use the following exponential function:
\begin{equation}
\label{eq: attractivenessFirefly}
\beta = \beta_{0}\ e^{-\gamma r_{ij}^2},
\end{equation}
where $r_{ij}$ is the distance defined as in Eq.~(\ref{eq: EuclideamDistance}), $\beta_{0}$ is the initial attractiveness at the distance
$r_{ij}$ $=0$, and $\gamma$ is an absorption coefficient at the source which controls the decrease of the light intensity.
The movement of a firefly $i$  which is attracted by a more attractive (i.e., brighter) firefly $j$ is governed by the following evolution equation:
\begin{equation}
\label{eq: movementoffirefly}
x_i^{t+1} = x_i^t+  \beta_{0}\ e^{-\gamma r_{ij}^2} ( x_j^t - x_i^t ) + \alpha(\sigma-\frac{1}{2}),
\end{equation}
where the first term on the right-hand side is the current position of the firefly $i$, the second term is used for modelling the attractiveness of the firefly as the light intensity seen by adjacent fireflies, and the third term is randomization with  $\alpha$ being the randomization parameter and it is determined by the problem of interest.
Though $\sigma$ was originally a random variable, we use it here as a scaling factor that controls the distance of visibility and in most case we can use  $\sigma \in [0,1]$.

Broadly speaking, when a robot $k$ detects a target, it switches to a Coordinator State, and it acts as a firefly, sending out help requests to its neighborhood $LN_k^t$. When a robot receives this request and it decides to contribute in the disarming process, it stores the request in its list $RR_K$. If the list contains more requests, it must choose which target it will disarm. Using the relative position information of the found targets, the robot derives the distance between it and the coordinators and then uses this metric to choose the best target, that is usually the closer. The same information also allows to derive the next movement of the robots. The approach provides a flexible way to decide when it is necessary to reconsider decisions and how to choose among different targets.

It should be noticed that the recruited robots do not respond to the received requests, since they can change their decision at any time, so the coordinators robots do not know which robots are recruiting and continue to broadcast packets until the needed robots have arrived. This has some implications. First, not all recruited robots will go towards the target's locations balancing the two task. Second, the order on which the requests are received is not as important as the allocation is not instantaneous. This allows an effective approach to reach solutions that the greedy strategy would miss. Third, the reduction of the impact on communications, so that bandwidth used will increase slowly with the team size.

\begin{figure}
	\flushleft
	\includegraphics[scale=0.32]{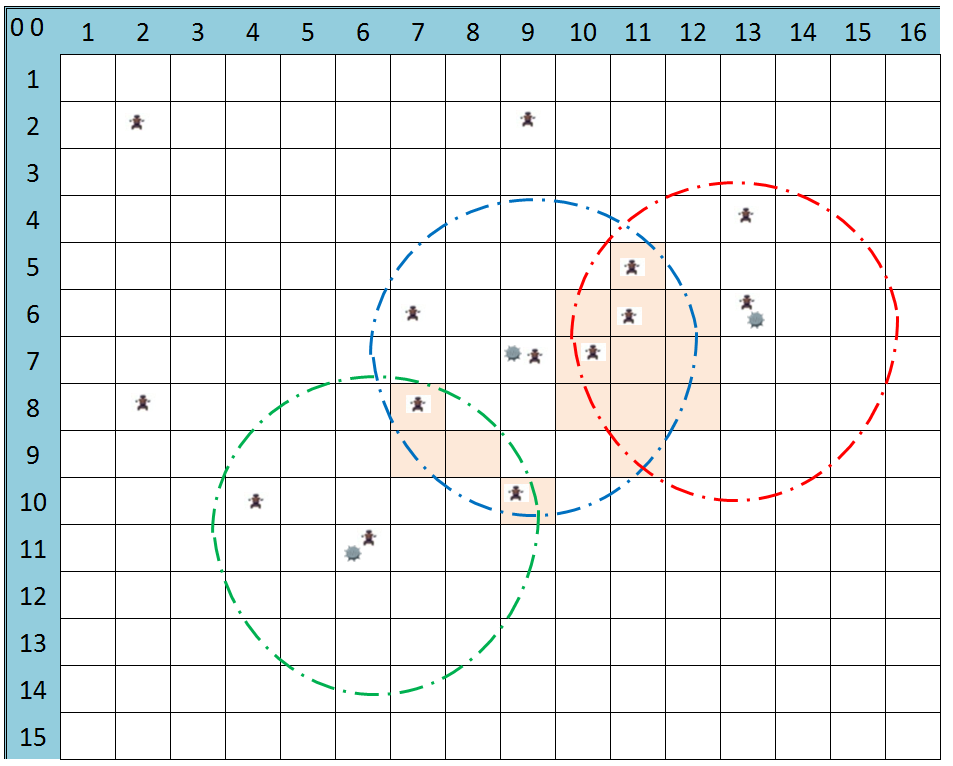}
	\caption{Example of an overlap region  in which some robots are in the wireless ranges of different coordinator robots and thus they must decide towards which target to move, according to the Firefly-based strategy. }
	\label{fig:OverlappingArea}
\end{figure}

Then the robots move towards target's location according to a modified version of the firefly algorithm. The aim of this strategy is to increase the flexibility of the system that let the robots be able to form groups effectively and efficiently in order to enhance the parallelism of the handling of the found targets, and at the same time move towards the target location's avoiding overlapping regions and any redundancy (Fig.~\ref{fig:OverlappingArea}).

Moreover, the algorithm allows to dynamically adjust the coordination task since it allows for each robot to make the best choice from its own point of view.

\subsubsection{Implementation of Robot Decision Mechanism}

The original version of FA is applied in the continuous space, and cannot be applied directly to tackle discrete problems, so we have modified the algorithm in order to solve our problem.  In our case, a robot can move in a 2-D discrete space and it can go just to the adjacent cells. This means that when a robot $k$, at iteration $t$, in the cell $c_k^t$ with coordinates $(x_k^t,y_k^t)$ receives a packet by a coordinator robot that has found a target, this robot will move in the next step $(t+1)$ to a new position $( x_k^{t+1},y_k^{t+1})$, according to the FA attraction rules such as expressed below:
\begin{equation}
\label{eq: movementoffireflyinyxdirection}
\begin{cases}
x_{k}^{t+1} = x_{k}^t+  \beta_{0}\ e^{-\gamma r_{kz}^2} (x_z - x_{k}^t) + \alpha(\sigma-\frac{1}{2}),
\\
\\
y_{k}^{t+1} = y_{k}^t+  \beta_{0}\ e^{-\gamma r_{kz}^2} (y_z - y_{k}^t) + \alpha(\sigma-\frac{1}{2}),
\end{cases}
\end{equation}
where $x_z$ and $y_z$ represent the coordinates of the selected target translated in terms of row and column of the matrix area, $r_{kz}$ is the Euclidean distance between the  target $z$ and the recruited robot. It should be noticed that the targets are static and a robot can receive more than one request. In the latter case, it will choose to move towards the brighter target within the minimum distance from the target as expressed in
Eq.~(\ref{eq: attractivenessFirefly}). A robot's movement is conditioned by the target's position and by a random component that it is useful to avoid the situation that more recruited robots go towards the same target if more targets have found. This last condition enables to the algorithm to potentially jump out of any local optimum (Fig.~\ref{fig:OverlappingArea}).

A key aspect occurs when robot $k$, moves too far from the target's position. Given a robot $k$ located at the step $t$ in the cell of coordinates $(x_k^t, y_k^t)$ and the target $z$ with coordinates $(x_z,y_z)$, we define the distance between the robot $k$ and the target $z$ as the Euclidean distance $$r_{kz} = \sqrt{(x_k^t-x_z)^2+(y_k^t-y_z)^2}. $$ If $r_{kz}$ $\geq$ $(R_t + \Delta)$ $\forall$ $z$ $\in$ $RR_k$  means that the robot $k$ moves too far from the target's locations and in this case, if it has not got other requests, it switches its role into Forager State (see Fig.~\ref{fig:ChangeState}).

\begin{figure}[t!]
	\flushleft
	\includegraphics[scale=0.37]{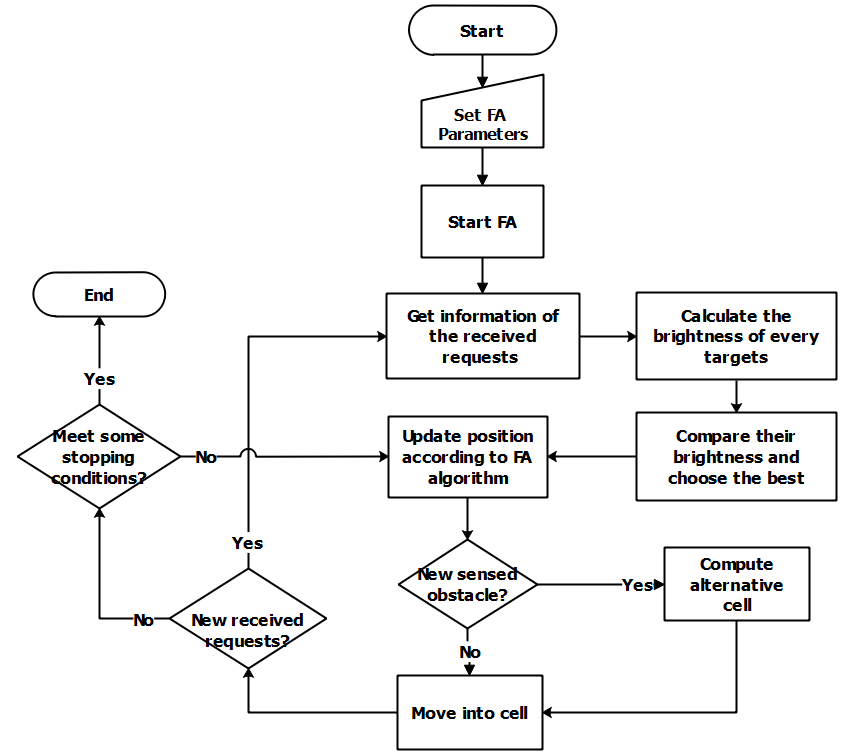}
	\caption{The flow chart of the Firefly Algorithm executed by a robot. At each step, the robots make decisions on the basis of events that can occur.}
	\label{fig:FAFlowChart}
\end{figure}

Fig.~\ref{fig:FAFlowChart} summarizes the main idea of the  FA-based decision mechanism.

In order to modify the FA to a discrete version, the robot movements have been modelled by three kinds of possible value updates for coordinates $\{$ -1, 0, 1 $\}$, according to the following conditions:

\begin{equation}
\label{eq: movementodiscretffireflyinXdirection}
\begin{cases}
x_{k}^{t+1} = x_{k}^t+ 1 & \text  { if [$\beta_{0} e^{-\gamma r_{kz}^2}     (x_z - x_{k}^t) + \alpha(\sigma-\frac{1}{2})  $   $>$ 0   ]},
\\ \\
x_{k}^{t+1} = x_{k}^t- 1 & \text  { if [$\beta_{0} e^{-\gamma r_{kz}^2}     (x_z - x_{k}^t) + \alpha(\sigma-\frac{1}{2})  $   $<$ 0   ]},
\\ \\
x_{k}^{t+1} = x_{k}^t & \text  { if [$\beta_{0} e^{-\gamma r_{kz}^2}     (x_z - x_{k}^t) + \alpha(\sigma-\frac{1}{2})  $   $=$ 0   ]},
\end{cases}
\end{equation}
and
\begin{equation}
\label{eq: movementodiscretffireflyinYdirection}
\begin{cases}
y_{k}^{t+1} = y_{k}^t+ 1 & \text  { if [$\beta_{0} e^{-\gamma r_{kz}^2}     (y_z - y_{k}^t) + \alpha(\sigma-\frac{1}{2})  $   $>$ 0   ]},
\\
\\
y_{k}^{t+1} = y_{k}^t- 1 & \text  { if [$\beta_{0} e^{-\gamma r_{kz}^2}     (y_z - y_{k}^t) + \alpha(\sigma-\frac{1}{2})  $   $<$ 0   ]},
\\
\\
y_{k}^{t+1} = y_{k}^t & \text  { if [$\beta_{0} e^{-\gamma r_{kz}^2}     (y_z - y_{k}^t) + \alpha(\sigma-\frac{1}{2})  $   $=$ 0   ]}.
\end{cases}
\end{equation}

\begin{figure}
	\centering
	\includegraphics[scale=0.4]{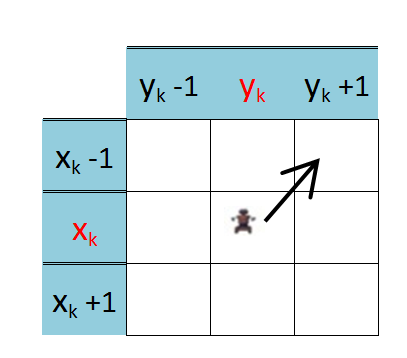}
	\caption{A possible selected cell after the application of a bio-inspired strategy.}
	\label{fig:MovementRobot}
\end{figure}

A robot (e.g., robot $k$) that is in the cell with coordinates $(x_k^t ,y_k^t)$ as depicted in  Fig.~\ref{fig:MovementRobot}  can move into eight possible cells according to the three possible values attributed to $x_k$ and $y_k$. For example, if the result of Eqs. (\ref*{eq: movementodiscretffireflyinXdirection})-(\ref*{eq: movementodiscretffireflyinYdirection}) is (-1, 1), the robot will move into the cell $(x_{k}^t-1, y_k^t+1)$.
In the described problem, the algorithm for the Firefly based strategy is shown in
Algorithm~\ref{FaAlgorithm}.

\begin{tiny}
	\begin{algorithm}[t!]
		\caption{Firefly based strategy.  }\label{FaAlgorithm}
		\SetAlgoLongEnd
		 \SetKwData{Left}{left}\SetKwData{This}{this}\SetKwData{Up}{up}
		\SetKwFunction{Union}{Union}\SetKwFunction{FindCompress}{FindCompress}
		\Begin {
			\SetKwInOut{Stepi}{Step 1}
			\SetKwInOut{Stepii}{Step 2}
			\SetKwInOut{Stepiii}{Step 3}
			\SetKwInOut{Stepiiii}{Step 4}
			
			\Stepi{ \textbf{Initialization.}
				Set $t$ $\{$t is the time step$\}$;\\
				Set the detected targets; \\
				Set the roboto in Recruited State; \\
				Define the light absorption \\coefficient $\gamma$;\\ Set the randomization parameter \\$\alpha$;\\ Set the random number $\sigma$; \\ Set the attractiveness $\beta_0$;}
	 \Stepii{\textbf{Generation coordination system.} \\For the detected targets and the\\ recruited robots, set the initial\\ locations in terms of coordinates \\ in $x$ and $y$ directions; }
			 \Stepiii{\textbf{Procedure.}}
			\While{The stop criteria are not satisfied}{
				\ForEach{robot $k$ in Recruited State }{\label{forins}
					set $RR_k$\;
					evaluate the current position $c_k^t$\;
					\ForEach{target $z$ $\in$ $RR_k$} {\label{forins}
						evaluate $\beta$ according to Eq. (\ref{eq: attractivenessFirefly})\;
						choose the best target $z$ \;}
					{evaluate $N(c_k^t)$\;
						compute the cell $c_k^{t+1}$ according to Eqs.(\ref{eq: movementodiscretffireflyinXdirection})-(\ref{eq: movementodiscretffireflyinYdirection})\;
						\eIf{($c_k^{t+1}$.hasObstacle() or  \hspace*{0.13cm} $c_k^{t+1}$.isOccupated() or \hspace*{0.13cm} $c_k^{t+1}$.isInaccessible())}
						{choose a random cell $c^*$ $\in$ $N(c_k^t)$;  \\
							move robot $k$ towards $c^*$;}
						{move robot $k$ towards $c_k^{t+1}$;}
					}
				}
				update $t$;
			}
			
		}
	\end{algorithm}
\end{tiny}

The Algorithm 2 is executed when one or more targets are found and some robots are recruited by others. If no target are detected or all targets are removed or handled, the robots perform the exploration task according to Algorithm~\ref*{ExplorationAlgorithm}.
More specifically, each recruited robot has the list of the requests in terms of target's locations and evaluates the brightness of each of them encoded as fireflies taking into account their distances.  At each step, the robots select the best from their list which has the maximum brightness. Next they move to the target's location according to Firefly-based rules.

The proposed firefly-based approach is computationally simple. It requires only few simple calculations (e.g., additions/subtractions) to update the positions of the robots. Moreover, the volume of information that is communicated among the robots is small, since only the position of the targets's is sent. For this reason, FTS-RR has the benefit of the scalability.
In addition, the algorithm tries to form a coalition with the minimum size of involved robots, so the remaining robots are able to conduct other search or disarmament tasks, allowing multiple actions at a time.

\subsection{Advantages}
A detailed analysis of the above proposed approach, we can highlight that
this hybrid approach combining indirect and direct communication has the following advantages:
\begin{itemize}
	\item It allows to dynamically adjust the importance of the various objectives used in order to get the best taken decision during each algorithm iteration.
	\item It introduces the concept of the importance of the task in terms of the weighted objective function, enabling robots to balance the two objectives when necessary.
	\item One of the most significant advantages of the approach is the flexibility of change decision at any time.
	\item It has a low computation cost since it is not required to know the decisions of the other robots, but each robot acts selfishly taking the best decision from its own point of view.
	\item Since the algorithm for the coordination is not constructed for the specific target type, all of them are treated in the exact same manner: the same variable types are used, regardless of what the target is and what the robot is performing. Therefore, the approach is generalized and can be used for a wide range of applications with minor modifications.
	\item Although each independent task is executed individually, the whole system can attempt to globally optimize the process.
	\item The system can adapt to changes in the environment, and thus the algorithm is scalable.
\end{itemize}

\section{Computational Experiments}

A set of experiments has been performed in order to show and analyze the effects of the weights on the performance of the swarm of robots. Simulation results have been summarized using different values of $w_1$ and $w_2$ in order to find the best configurations to balance the two tasks. More specifically, we have evaluated the values of the weights under different conditions and by varying the parameters of the system.
A simulator has been implemented in Java and for each set of experiments, the simulations have been repeated 50 times, thus the presented results are the mean values of those iterations. Each experiment starts with a random configuration of robots and targets.

\subsection{Test Parameters and Metrics}

An important aspect in evaluating search and rescue tasks is the definition of effective metrics for measuring the performance of the swarm. The performance metrics, in this paper, can be divided into two groups: measure of the time or energy needed to acquire all the information of the environments and the measure of information acquired with limited resources in terms of time, energy or other constraints.

The first case considers a static scenario where it is supposed that the robots have enough resources to explore the area and disarm all disseminated targets. To measure the performance, we have used two metrics that are the total time steps to complete the mission and the total energy consumed by the robots. For the purposes of calculating this cost metric, some the parameters and values are summarized in Tables \ref*{tab: explorationParameters}-\ref*{tab: WirelessCost}-\ref{tab: CoordinationParameters} (see \cite{42},\cite{44}).

\begin{table}
	\caption{Parameters used in the exploration algorithm. }
	\label{tab: explorationParameters}
	\begin{tabular}{cc}
		\hline\noalign{\smallskip}
		Parameters & Value  \\
		\noalign{\smallskip}\hline\noalign{\smallskip}
		Sensing range $R_s$  & 4 \\
		$\rho$ & 0.1  \\
		$\Delta\tau_0$ & 2 \\
		$\varphi$ & 1 \\
		$\lambda$ & 1 \\
		$\eta$ & 0.9\\
		$a_1$ & 0.5 \\
		$a_2$ & 0.5\\
		$\varepsilon$ & Uniform [0 1]\\
		\noalign{\smallskip}\hline
	\end{tabular}
\end{table}


In the model, $e_{cc}$, $e_{tx}$ and $e_{rc}$ have been recalculated to express them in terms of the unit of energy.


\begin{table}
	\caption{Parameters used in the Firefly Algorithm.}
	\label{tab: CoordinationParameters}
	\begin{tabular}{cc}
		\hline\noalign{\smallskip}
		Parameters & Value  \\
		\noalign{\smallskip}\hline\noalign{\smallskip}
		$\alpha$ & 0.2 \\
		$\beta_0$ & 0.5 \\
		$\gamma$ & $\frac{1}{L}$ ($L$=max$\{$m,n$\}$) \\
		$\sigma$ & Uniform [0,1] \\
		\noalign{\smallskip}\hline
	\end{tabular}
\end{table}

\begin{table}
	\caption{Cost related to the wireless communication.}
	\label{tab: WirelessCost}
	{\small
		\begin{tabular}{lc}
			\hline\noalign{\smallskip}
			Parameters & Value  \\
			\noalign{\smallskip}\hline\noalign{\smallskip}
			Bit Rate (B)   & 3 \\
			Energy Consumed by a transceiver circuit to  \\
			transmit or receive a bit, $e_{cc}$ (Joule) & $10^{-7}$ \\
			Energy Consumed by a transceiver amplifier to \\
			transmit 1 bit data per meter, $e_{tx}$ (Joule) & $10^{-12}$ \\
			Energy to receive a bit, $e_{rc}$ (Joule) & $10^{-7}$ \\
			Path loss Exponent, $\psi$ & [2,6] \\
			Wireless Range $R_t$ & $6,15$\\
			\noalign{\smallskip}\hline
		\end{tabular}
	}
\end{table}

At each step of the simulation, a robot will consume an amount of energy varying its state since the robots employ different actions in different states.
For example, a robot will consume more energy when handling a target than when wandering in the search area. A robot consumes 1 unit of energy for traveling from one cell to another. One stop takes an extra energy of 0.5 units. A turn of $45^{\circ}$ takes  0.4 units of energy. Turns of $90^{\circ}$, $135^{\circ}$, $180^{\circ}$, take 0.6, 0.8 and 1 units of energy, respectively. These numbers are approximately derived from energy measurements for Pioneer 3-DX robot \cite{45}. We estimate the energy for performing a planned task for removing or dismantling a target is about  5 units of energy for each robot involved in the task.

The second case considers a dynamic scenario, where a complete discovery of the information from the environment could be not feasible, due to resource constraints and unpredictable events. In this case, the performance metric is given by a set of functions that measure the percentage of information related to particular objectives. For the exploration task, the metric of interested is related to the amount of unexplored map in terms of unvisited cells. While, for the case of disarmament task, the metric takes into account the percentage of detected and disarmed targets/mines. Moreover, it is considered the number of alive robots at the end of the simulations that are the robots that have not finished their budget of energy and not exploded for external events.

Regarding the simulations, there are several test-related parameters that may influence the performance and the results and they are listed as follows:
\begin{itemize}
	\item Area size:  We study both scenarios with and without obstacles. In this study, we use square areas for their symmetry and simplicity. In the future, we may experiment with non–square areas, or areas with no boundaries at all. (The presence of boundaries makes the problem simpler by focusing the robots on
	the area of interest; in a world with no boundaries, the robots could get permanently lost.)
	\item Robot density: This is the total  number of robots in the swarm $|N^R|$.
	\item The number of targets $|N^T|$.
	\item The number of coalitions that is the minimum number of robots that can handle properly a target $R_{min}$.
	\item The transmission range $R_t$, which can have an effect on the recruiting task
\end{itemize}

\subsection{Evaluation of the Weights under Static Conditions}

In this section it is assumed that the robots have sufficient resources in terms of energy to execute the mission and the targets are static without possibility of explosion.

\subsubsection{Case study 1}

In the first set of experiments, the influence of the weights on the dimension of the area is taken into account. We have considered 50x50 square cells, 100x100 square cells, varying the team size (25, 40, 50 robots) and the number of dispersed targets. All experiments were carried out using 3 robots needed to handle a target properly.

\begin{figure}[t!]
	\centering
	\subfloat[][\emph]
	{\includegraphics[width=.85\textwidth]{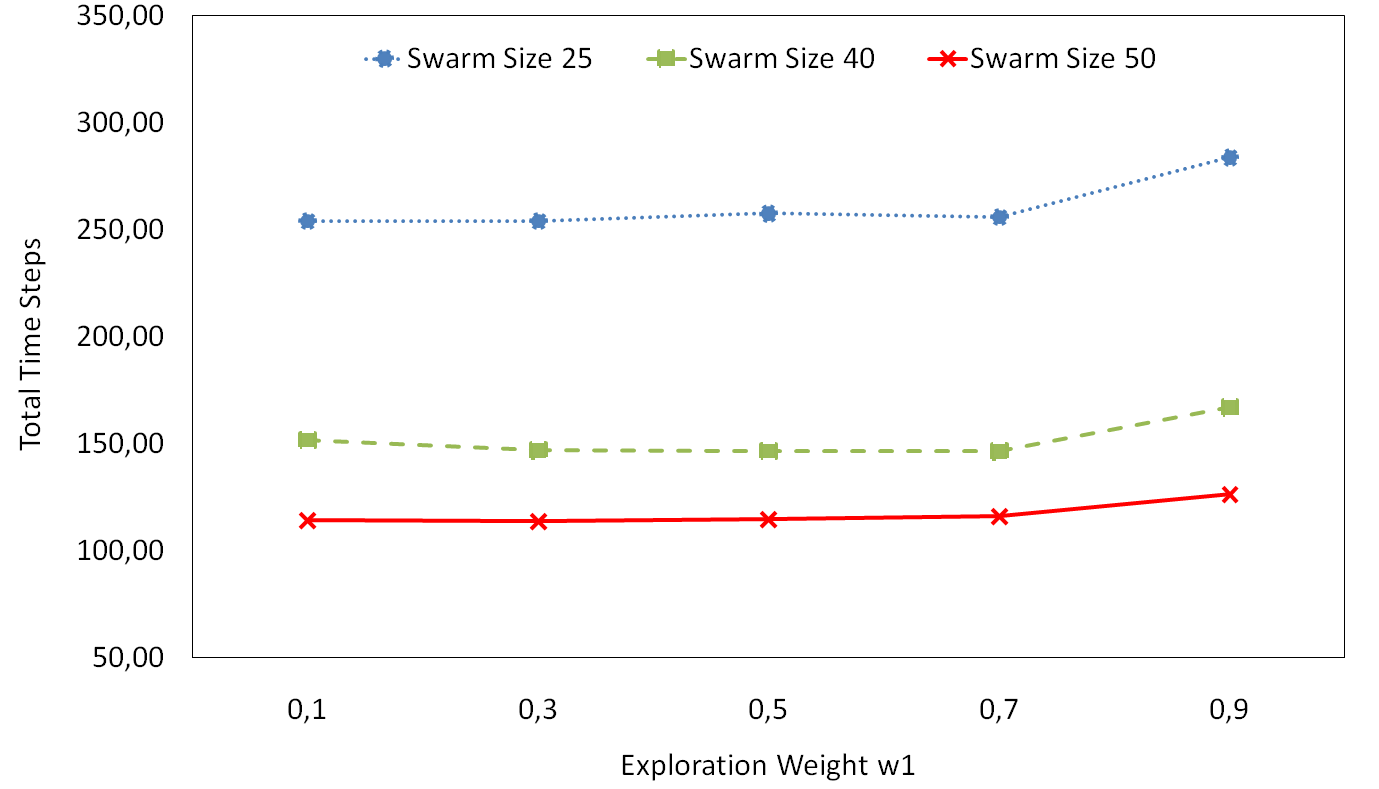}} \quad
	\subfloat[][\emph]
	{\includegraphics[width=.85\textwidth]{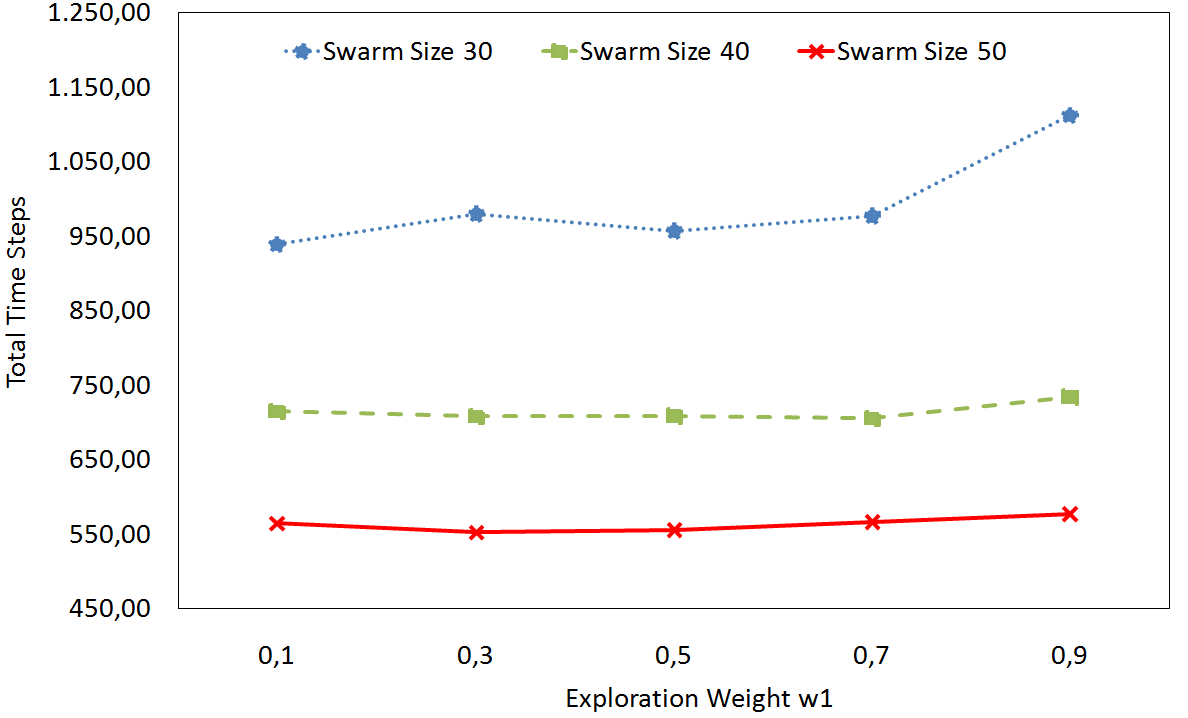}} \quad
	\caption{Evaluation of the Total Time Steps to complete the mission and 3 robots needed to handle a target. (a) 50x50 grid area, 10 targets to disarm  (b) 100x100 grid area, 20 targets to be disarmed. }
	\label{fig:DimensionInfluenceTime}
\end{figure}

\begin{figure}[t!]
	\centering
	\subfloat[][\emph]
	{\includegraphics[width=.85\textwidth]{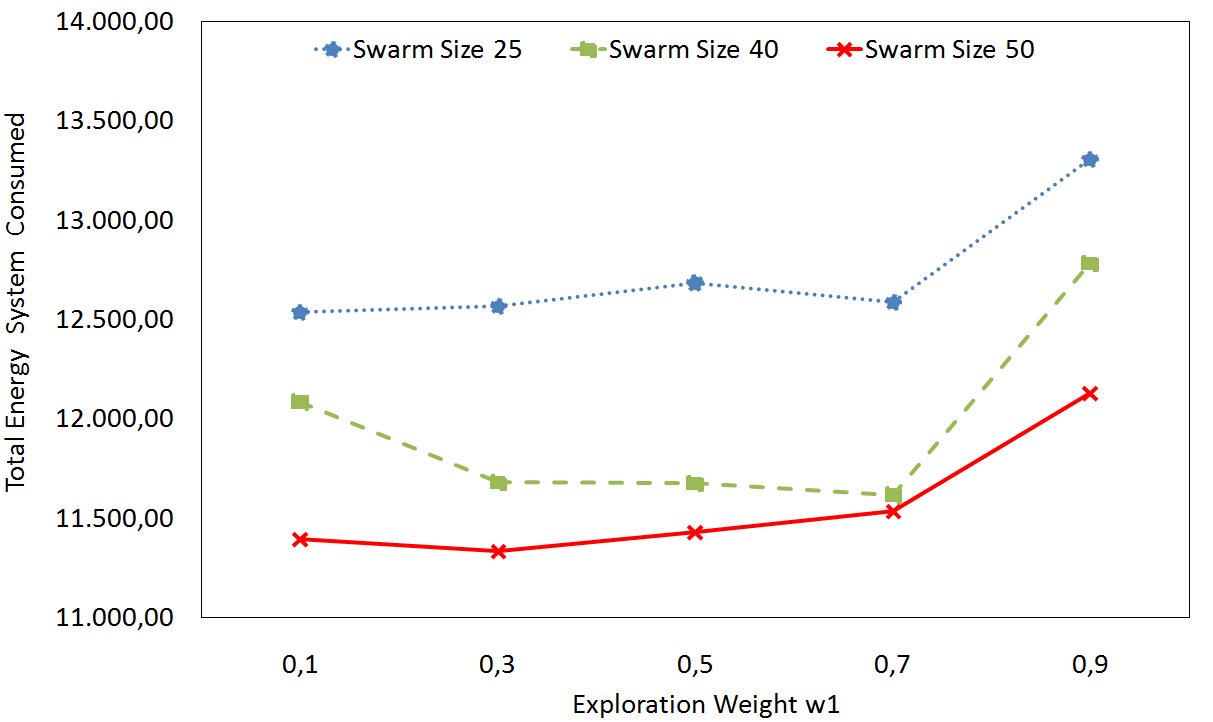}} \quad
	\subfloat[][\emph]
	{\includegraphics[width=.85\textwidth]{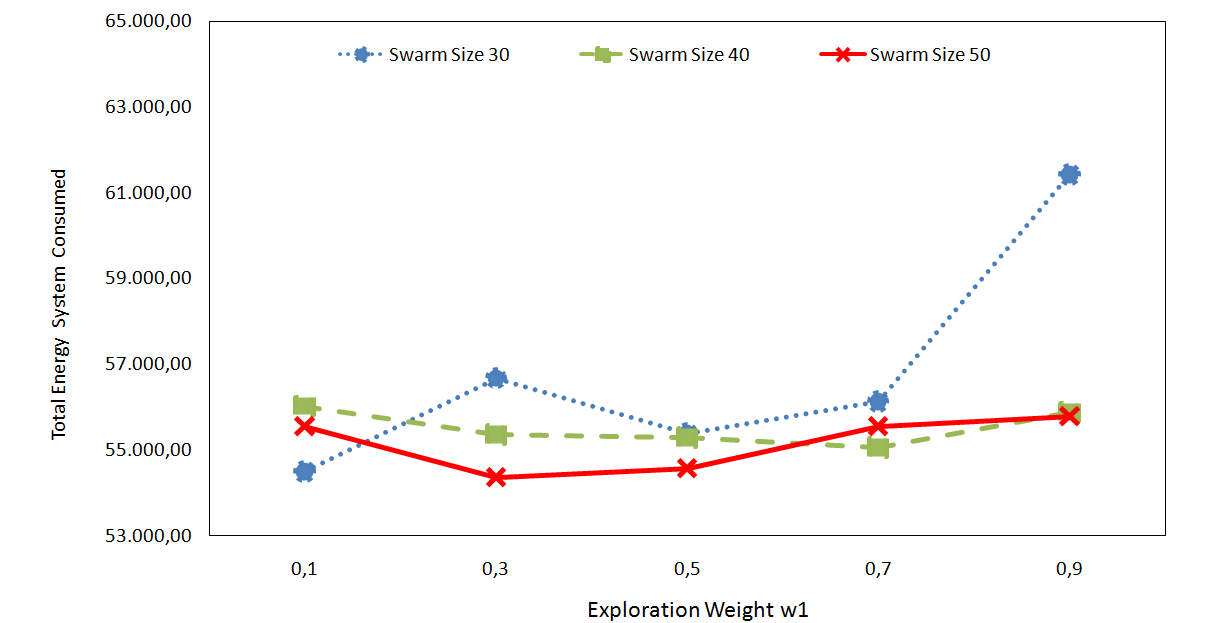}} \quad
	\caption{Evaluation of the Total-Energy-System- Consumed (TESC) (a) 50x50 grid area, 10 mines to disarm and 3 robots needed to handle a target. (b) 100x100 grid area, 20 targets to disarm and 3 robots needed to handle a target.}
	\label{fig:DimensionInfluenceEnergy}
\end{figure}

Figures \ref{fig:DimensionInfluenceTime} and \ref{fig:DimensionInfluenceEnergy} show the influence of the $w_1$ considering different swarm size and dimension of the area evaluating respectively the total time steps to complete a mission and the total energy consumed by the swarm. It can be observed that the time steps increase as the value $w_1$ increases when the size of the swarm is small. This behaviour can be explained, by observing the nature of the mission that implies the collaboration of more robots in target's locations. When $w_1$ increases, the robots are highly motivated to explore the area. Since, the mission is completed if all target are found, motivating the robots to explore the area than disarming targets, which can lead to a temporary deadlock, especially when the swarm size is small compared to the dimension of the area and the complexity of the mission in terms of the targets. This implies the decrease of the performance of the entire system.
On the other hand, a team with a larger number of robots generally increases the performance improvements. The curves do not fluctuate a lot and the total time steps is almost similar for different $w_1$ values. This implies that the influence of $w_1$ on the performance in general decreases, considering an adequate swarm size.

Regarding the energy consumption, it can be seen a high wastage of energy resource, considering the same total time steps, when  $ 0.7 \leq w_1 \leq  0.9 $. This difference is higher in teams with a low number of robots, compared to the number of disseminated targets  $\frac{N^R} {R_{min}*N^T}$ $\ll$ 1 , and in a big grid area (e.g., 30 robots operating in 100x100 cells for treating 20 targets).
Considering both the total time steps and the total energy consumption, for almost all cases the best range is $0.3 \leq w_1 \leq 0.5$.

\subsubsection{Case study 2}

\begin{figure}[t!]
	\flushleft
	\subfloat[][\emph]
	{\includegraphics[width=.85\textwidth]{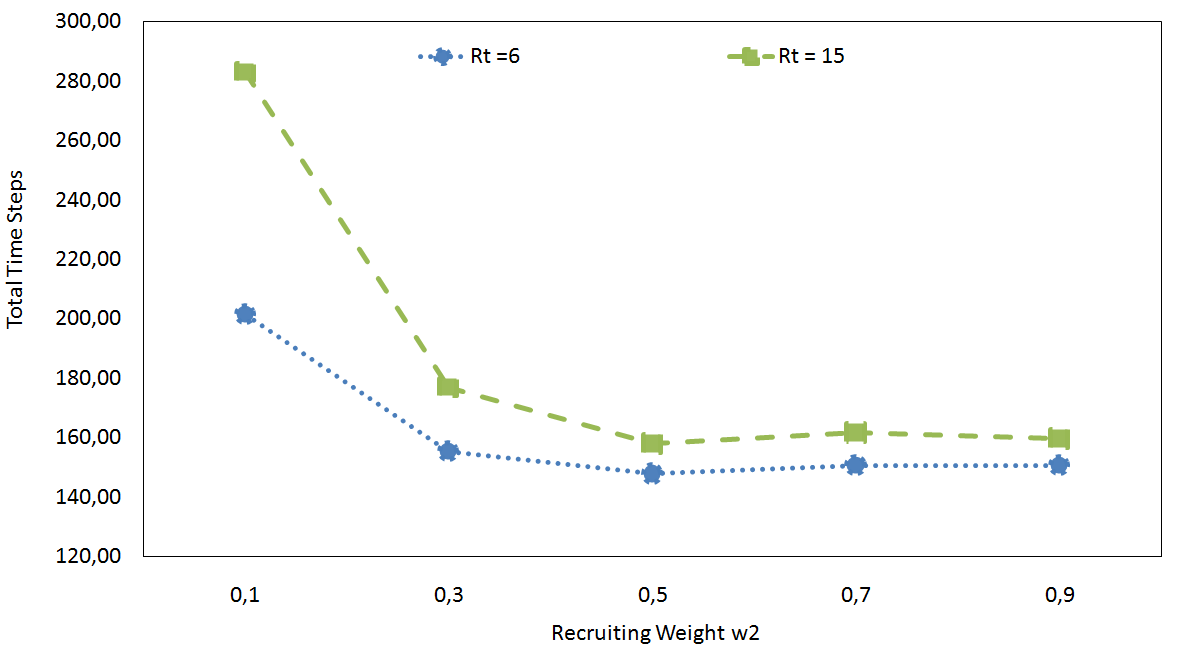}} \quad
	\subfloat[][\emph]
	{\includegraphics[width=.85\textwidth]{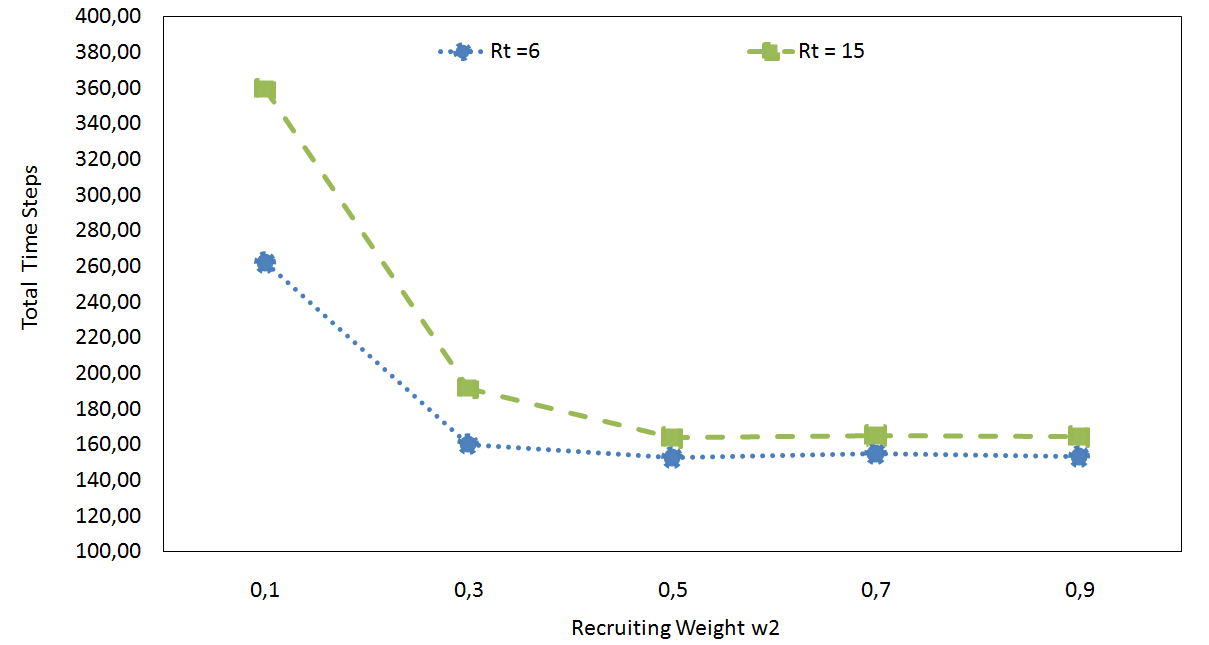}} \\
	\subfloat[][\emph]
	{\includegraphics[width=.85\textwidth]{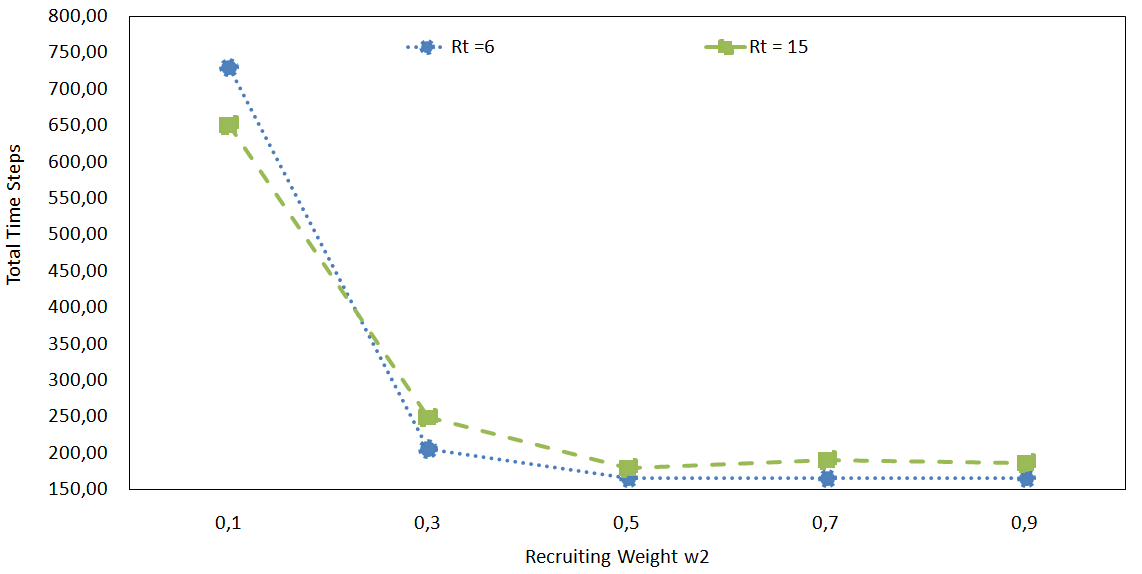}} \quad
	
	\caption{Evaluation of the total time steps to execute the mission in 50x50 squares and 40 robots and 3 robots to handle a target. (a) 15 targets to be handled. (b) 20 targets to be handled. (c) 35 targets to be handled. }
	\label{fig:TransmissionRangeTime}
\end{figure}

The second set of simulations compares the performance by varying the transmission range $R_t$ (6, 15 units of cells) considering a grid area 50x50 , a team with 40 robots and by varying the number of disseminated targets (15, 20, 35). This can play an important role in recruiting tasks, since for a higher transmission range, the probability that more robots are recruited increases. Figure~\ref{fig:TransmissionRangeTime} shows the total time steps under different conditions in terms of dispersed targets and the same swarm size (40 robots operating in the area).

It can be observed that an increase of the transmission range does not always imply the increase in performance in terms of time steps. The reason could be explained that if resources are enough in terms of robots as shown in Fig. \ref{fig:TransmissionRangeTime} (a)-(b), an increase of  the transmission range can cause some redundancy with the wastage of time to complete the mission. For example, considering a  small team compared to the targets (40 robots and 35 targets),
Fig.~\ref{fig:TransmissionRangeTime} (c) shows that a high transmission range with a small $w_2$ may imply a better performance. By increasing $w_2$, more robots may be involved in the recruitment task and there is no significant difference between the two ranges.
As expected, if the number of the targets to be handled is small, a high transmission range deteriorates the performance since unnecessary robots could be involved in the disarmament process, depleting resources for exploration task and eventually to other targets.
Although the total time steps are somewhat better, the performance in terms of energy consumed by the system strongly degrades, using a high transmission range.

\begin{figure}[t!]
	\flushleft
	\subfloat[][\emph]
	{\includegraphics[width=.85\textwidth]{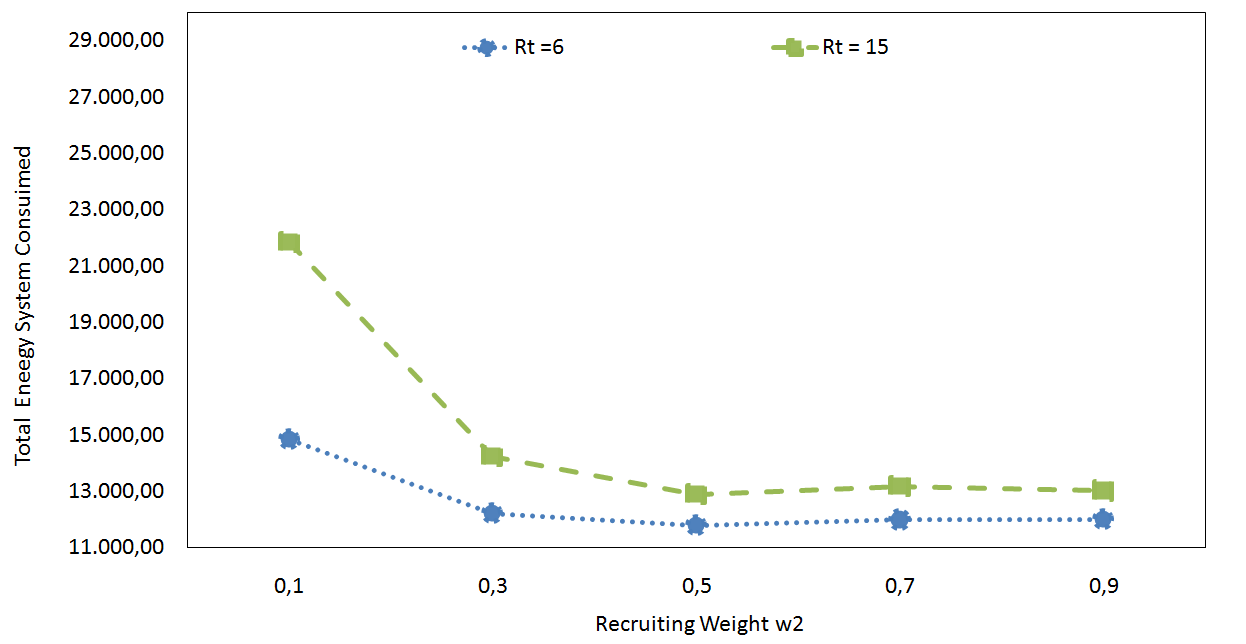}} \quad
	\subfloat[][\emph]
	{\includegraphics[width=.85\textwidth]{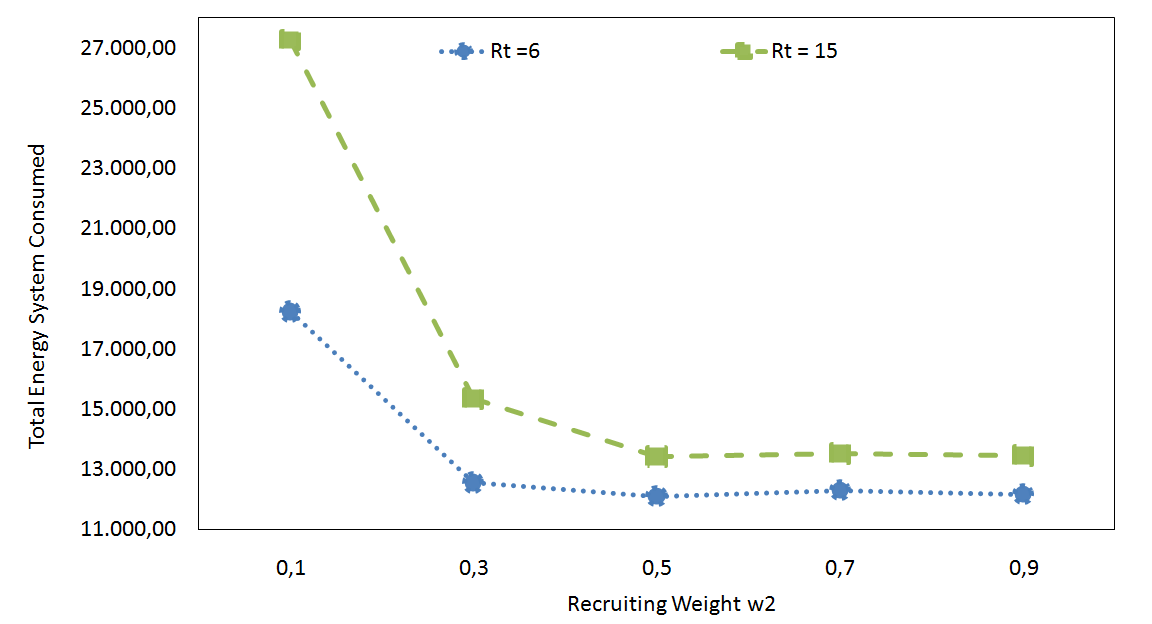}} \\
	\subfloat[][\emph]
	{\includegraphics[width=.85\textwidth]{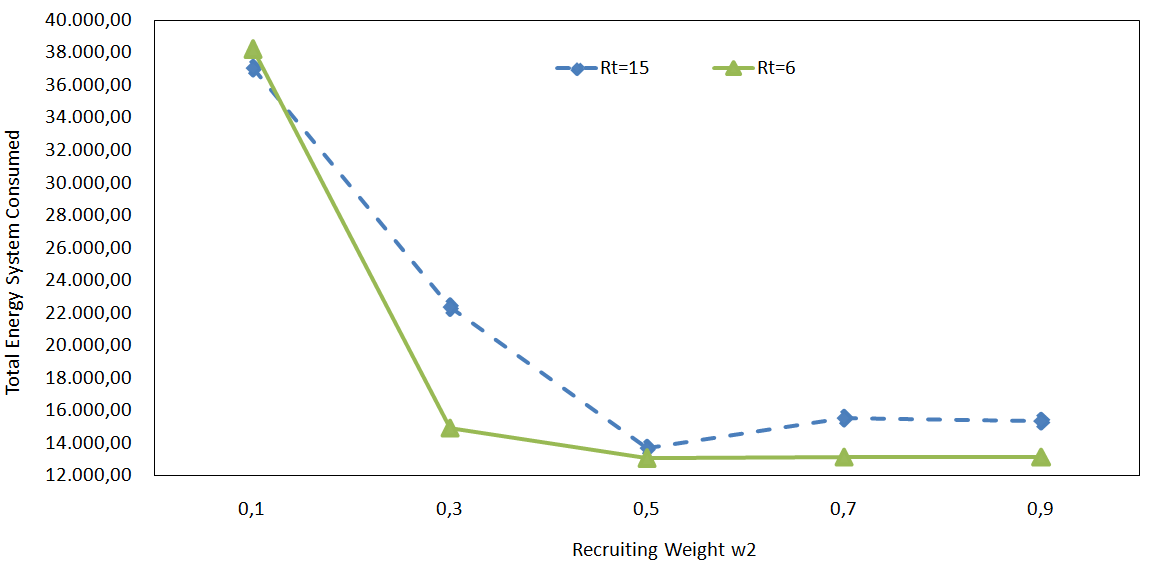}} \quad
	
	\caption{Evaluation of the total energy consumed by the system in 50x50 squares and 40 robots and 3 robots to treat a target. (a) 15 targets (b) 20 targets (c) 35 targets . }
	\label{fig:TransmissionRangeEnergy}
\end{figure}

Regarding the impact of $w_2$ on the performance, the effect of increasing the transmission range  can be quantitatively notated by looking at Fig.~\ref{fig:TransmissionRangeEnergy}. The results confirm that, especially for complex missions with $\frac{N^R} {R_{min}*N^T}$ $\ll$ 1,   more importance should be emphasized to the recruiting weight, thus  $w_2$ $\ge$ $0.3$.

\subsubsection{Case study 3}

The third set of simulations investigates the effect of the weights in relation to the number of disseminated targets. The performance measures have been evaluated by varying the dimension of the area, the swarm size and the number of robots that can deal with a target (2, 3, 4, 5).

\begin{figure}[t!]
	\flushleft
	\subfloat[][\emph]
	{\includegraphics[width=.85\textwidth]{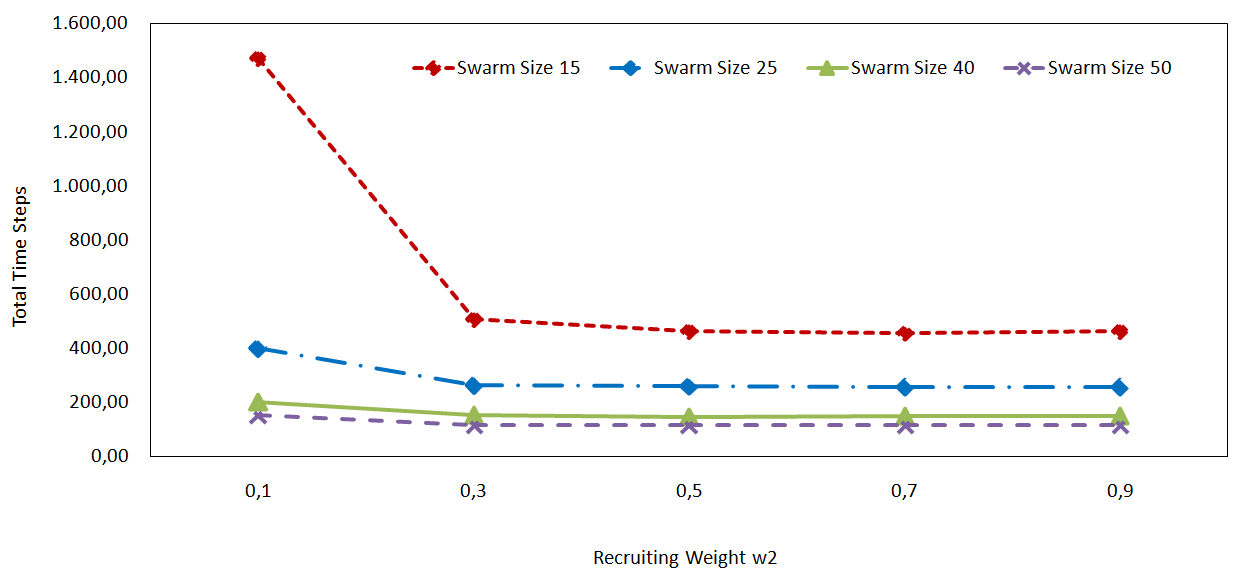}} \quad
	\subfloat[][\emph]
	{\includegraphics[width=.85\textwidth]{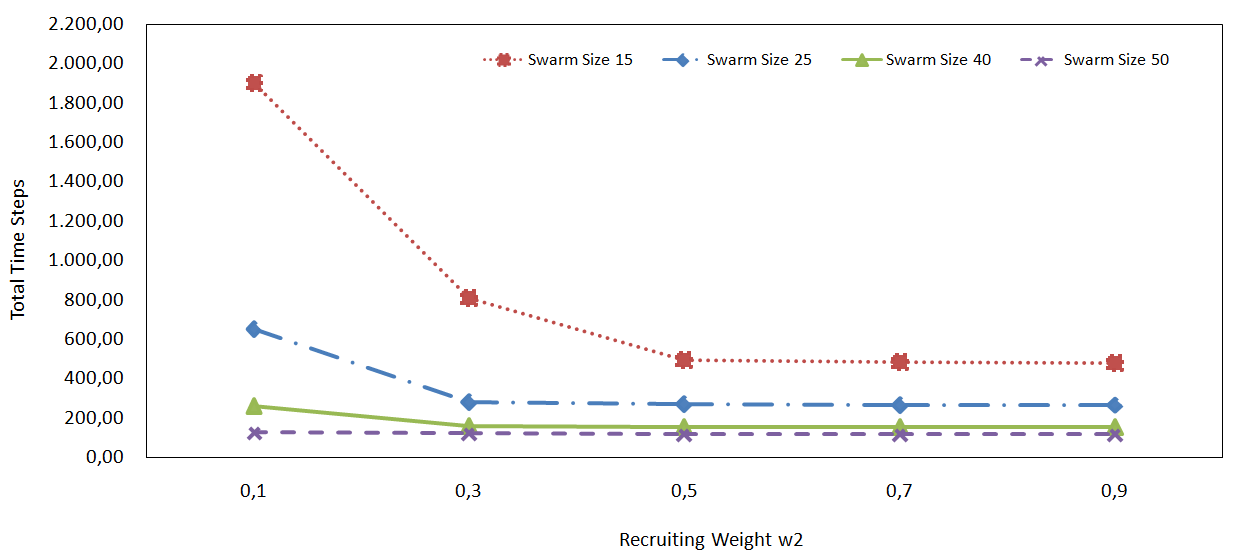}} \\
	\caption{Evaluation of the total time steps in a grid area 50x50. (a) 15 targets. (b) 20 targets. }
	\label{fig:50x50DifferentRobots}
\end{figure}

\begin{figure}[t!]
	\flushleft
	\subfloat[][\emph]
	{\includegraphics[width=.85\textwidth]{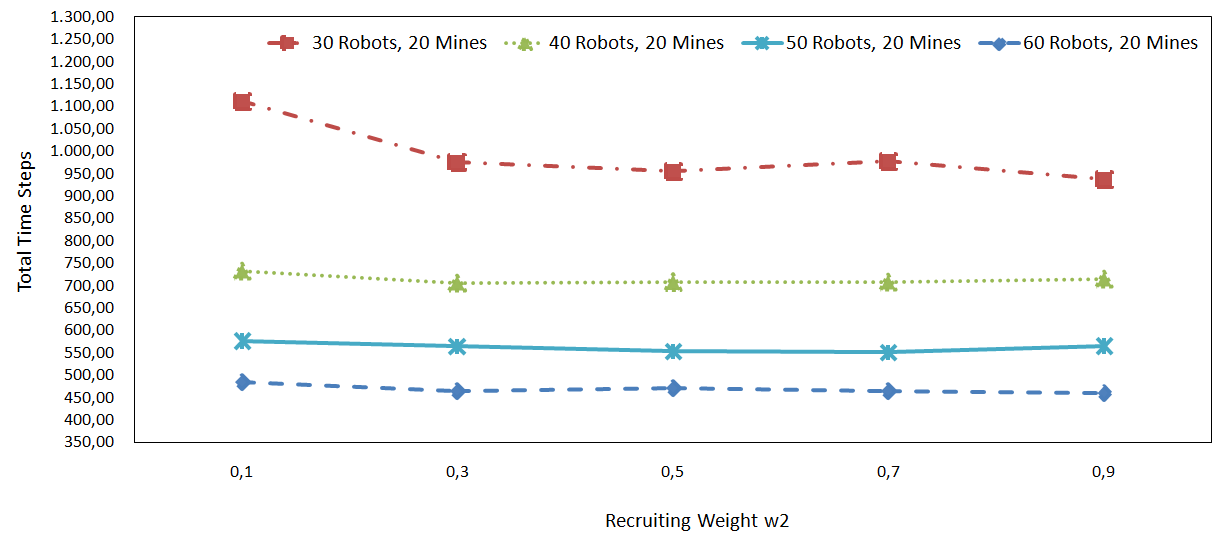}} \quad
	\subfloat[][\emph]
	{\includegraphics[width=.85\textwidth]{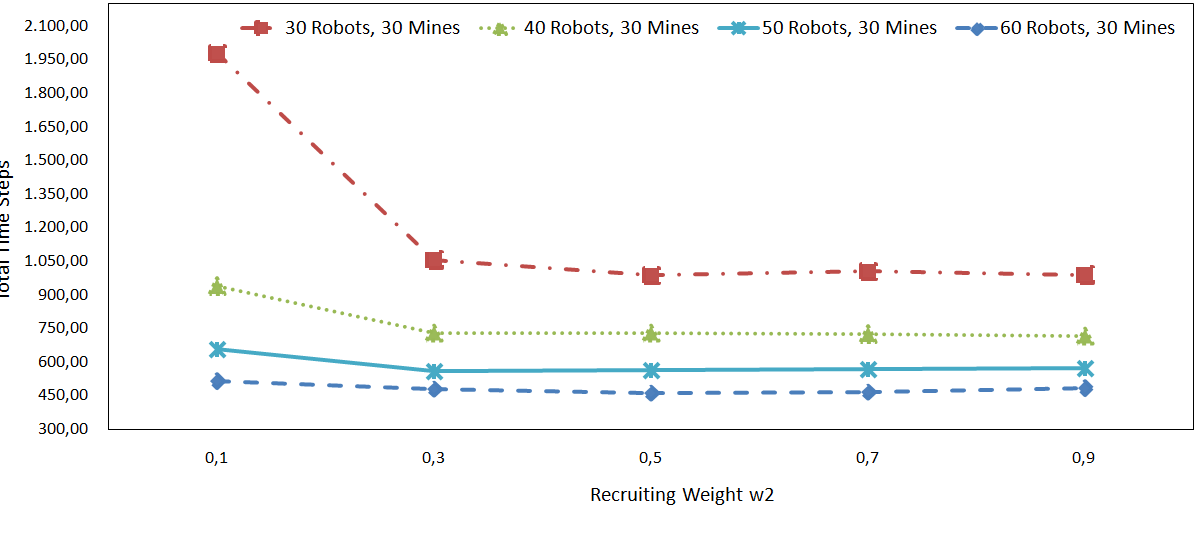}} \\
	\caption{Evaluation of the total time steps in a 100x100 grid area, varying the dimension of the swarm with (a) 20 targets and (b) 30 targets. }
	\label{fig:100X100DifferentRobots}
\end{figure}

The importance of choosing the $w_2$ weight properly increases as the number of the dispersed targets increases and it depends mostly on the dimension of the swarm. More specifically, if  $\frac{N^R} {R_{min}*N^T}$ $\ll$ 1 means the task is complex in terms of disarmament, more importance can be attributed to $w_2$ than $w_1$ as shown in Figures~(\ref{fig:50x50DifferentRobots})- (\ref{fig:100X100DifferentRobots}).

On the other hand, if  $\frac{N^R} {R_{min}*N^T}$ $\approx 1$, no significant influence in terms of total time steps is observed.
Obviously,  more robots are introduced, less wastage of time can be observed and $w_2$ becomes less relevant.

\begin{figure}[t!]
	\centering
	\subfloat[][\emph]
	{\includegraphics[width=.85\textwidth]{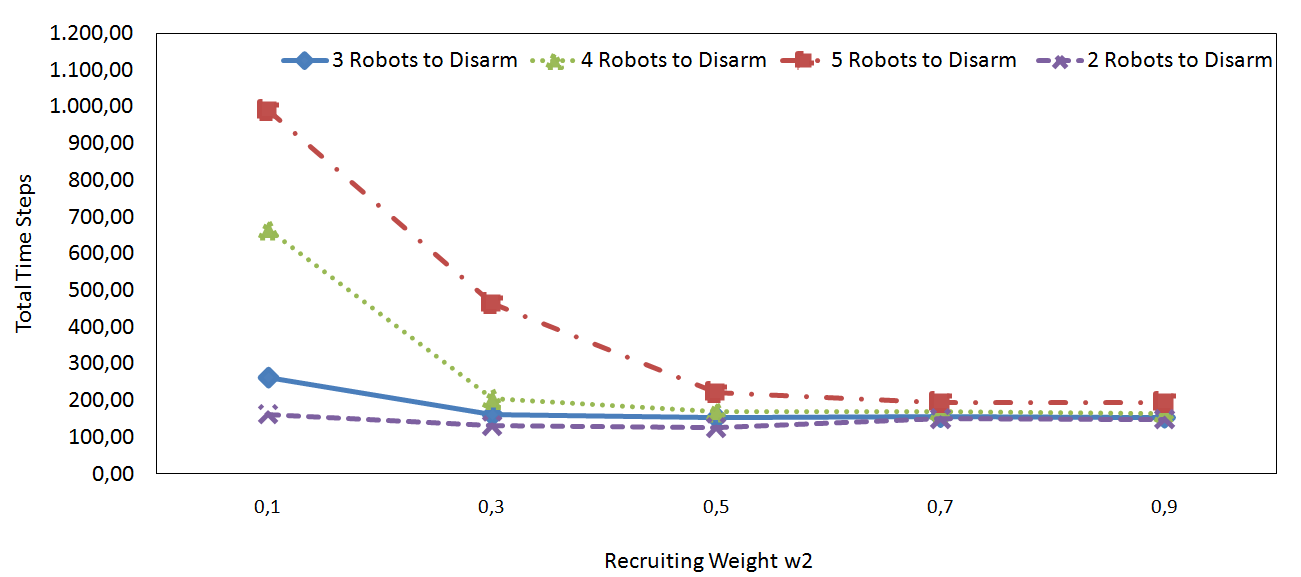}} \quad
	\caption{Evaluation of the total time steps varying $R_{min}$ in an area with 50x50 squares, 40 robots and 20 targets.}
	\label{fig:DifferentRoboToDisarm}
\end{figure}

In another set of experiments, we have introduced an additional parameter to control the task complexity; that is, the number of robots $R_{min}$ required for treating a target. In this way, we can vary the task complexity and observe its influence on the impact of $w_2$.
Fig.~\ref{fig:DifferentRoboToDisarm} shows the impact of $R_{min}$ and $w_2$ on the performance in terms of total time steps. It can be noticed that a high number of robots necessary to handle a target (5 robots to disarm) can cause severe resources consumption in terms of total time steps, if a small value is assigned to $w_2$. This leads to weakening the ability of the robots to distribute into a target's position. As a result, the robots wondering in the area increase the time to complete the mission and the coordinators can be trapped in target's location for a long time. Thus, increasing $R_{min}$, the $w_2$ weight should be increased in order to speed up the formation of the coalition. So in this case, $w_2$ can  greatly influence the performance and a proper value should be chosen ($w_2$ $\geq$ 0.5).
On the other hand, if the disarmament task is not particularly complex, the influence of the $w_2$ decreases.

\subsection{Evaluation of weights under dynamic conditions}

This section investigates the effect of the weights operating in a dynamic scenario where unpredictable events can occur (such as explosion of the mines and energy constraints).
It is assumed that a robot has 1000 energy units \cite{46}, without possibility of recharging during the mission, which means that if a robot consumes its energy, it will stop to perform the task at any time. In this case, to achieve good coordination and exploration is more challenging since it is required that the robot team has to respond quickly, robustly, reliably and adaptively to unexpected events.

To measure the performance of such a robot team in practice, we consider a number of metrics applicable to the performance of the individual robots and the team as a whole. More specifically, we consider the percentage of unexplored cells, the number of disarmed targets and the percentage of alive robots.

\begin{figure}[t!]
	\flushleft
	\subfloat[][\emph]
	{\includegraphics[width=.85\textwidth]{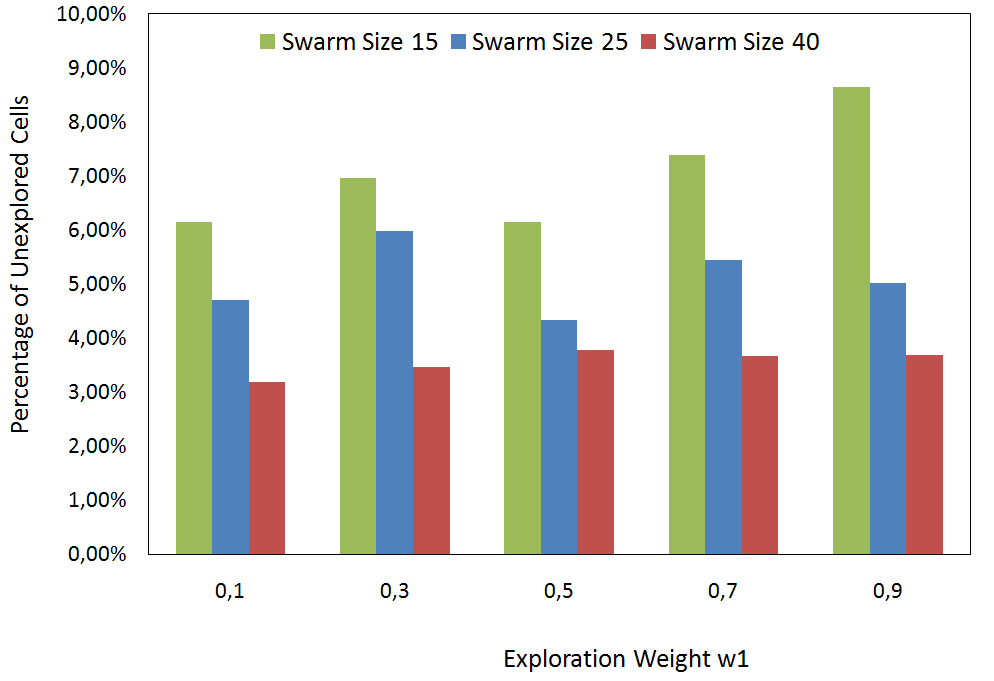}} \quad
	\subfloat[][\emph]
	{\includegraphics[width=.85\textwidth]{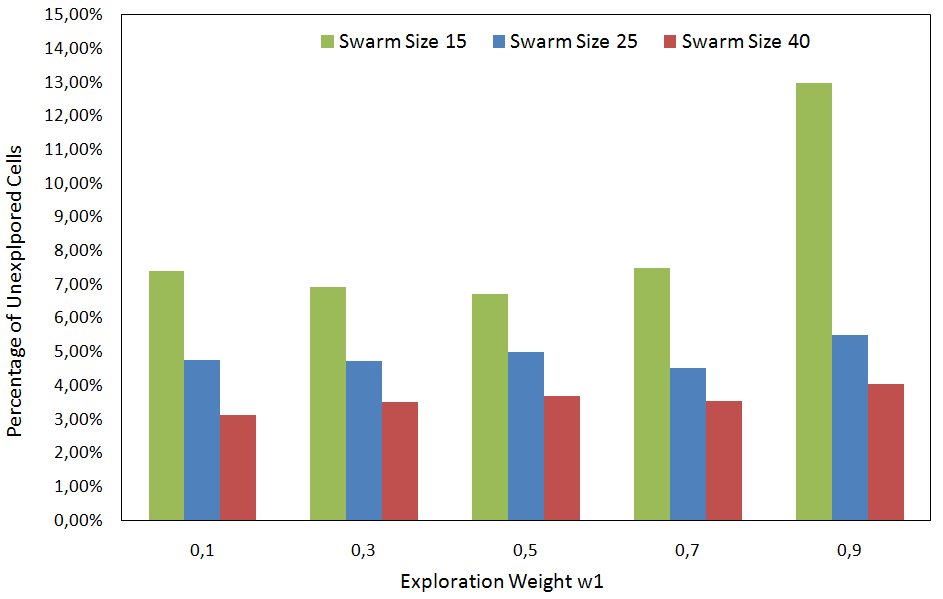}} \\
	\subfloat[][\emph]
	{\includegraphics[width=.85\textwidth]{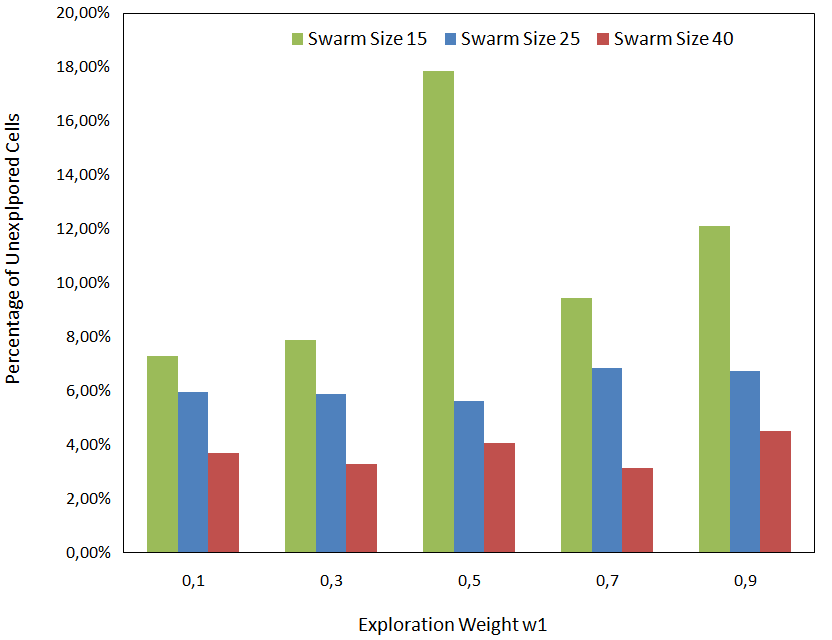}} \quad
	
	\caption{Percentage of unexplored cells in a 50x50 grid area, varying the dimension of the swarm of robots and 3 robots needed to handle a target. (a) 10 targets (b) 15 targets (c) 20 targets }
	\label{fig:Dynamic50x50UnexploredCells}
\end{figure}

Fig.~\ref{fig:Dynamic50x50UnexploredCells} shows the  impact of $w_1$ on the unexplored cells,  varying both the dimension of the swarm and the number of dispersed targets, while keeping $R_{min}$ as a constant. It can be noticed that for a small robot team and a hight number of targets, the performance degrades as $w_1$ increases.
This happens because the targets, for example mines, can explode at any time, causing not only the sudden stop of some robots in nearby regions, but also the damage of possible unexplored cells that become inaccessible. In these situations, the best value is about $w_1$ $\leq$ 0.5, which allows to balance the two tasks.

\begin{figure}[t!]
	\flushleft
	\subfloat[][\emph]
	{\includegraphics[width=.7\textwidth]{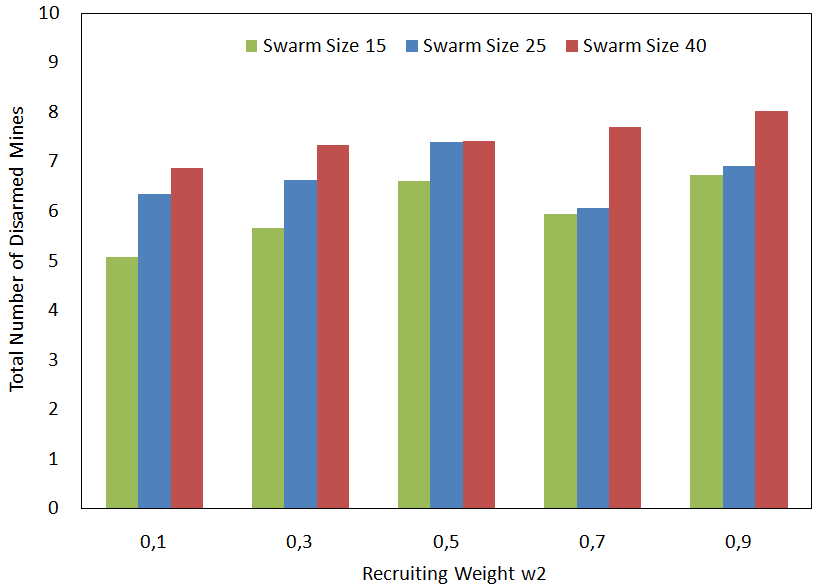}} \quad
	\subfloat[][\emph]
	{\includegraphics[width=.7\textwidth]{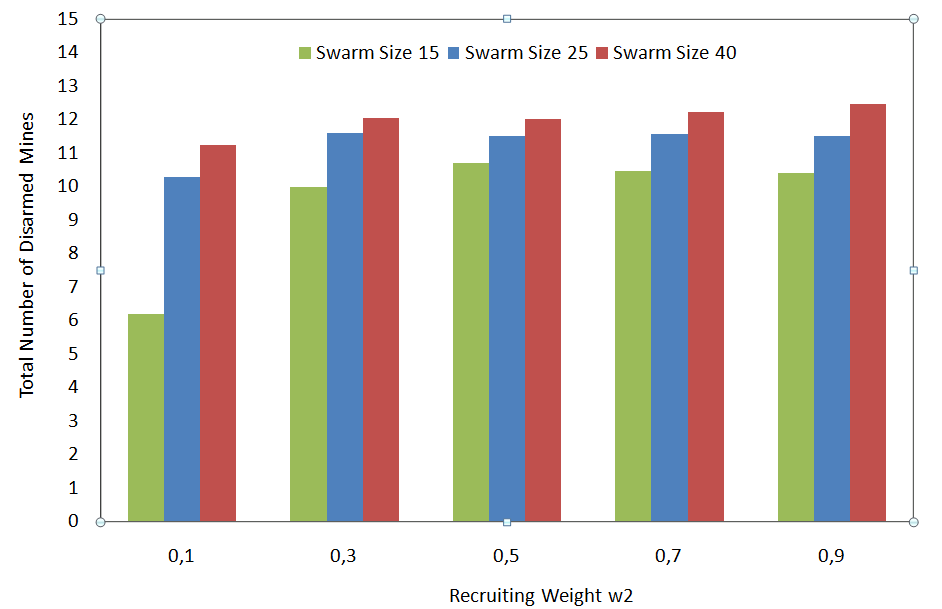}} \\
	\subfloat[][\emph]
	{\includegraphics[width=.7\textwidth]{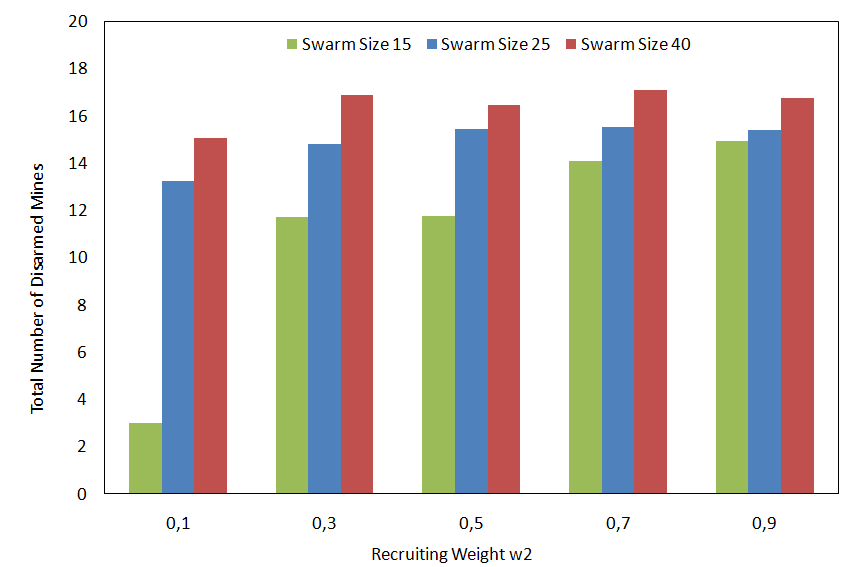}} \quad
	
	\caption{Percentage of disarmed targets in a 50x50 grid area, varying the dimension of the swarm of robots and 3 robots needed to handle a target. (a) 10 targets (b) 15 targets  (c) 20 targets }
	\label{fig:Dynamic50x50DisarmedMines}
\end{figure}

Regarding the disarmed targets, Fig.~\ref{fig:Dynamic50x50DisarmedMines} highlights the impact of $w_2$ on the number of disarmed targets. It can be noticed that the value is particularly important for small robot teams (15 robots and 20 targets to be disarmed) and  more importance can be attributed to the recruitment process ($w_2$ $\geq$ 0.3).
One possible explanation is that a robot team with higher movitation to be involved in the disarming process, can form a coalition more easily to handle a target so as to decrease the probability that it can explode.

\begin{figure}[t!]
	\flushleft
	\subfloat[][\emph]
	{\includegraphics[width=.6\textwidth]{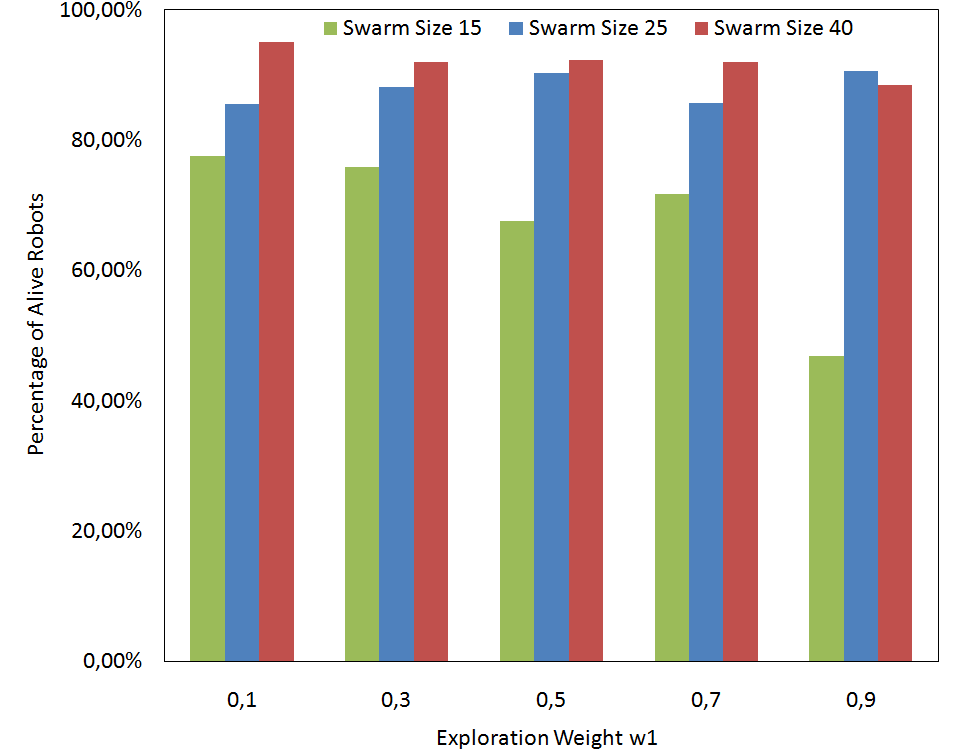}} \quad
	\subfloat[][\emph]
	{\includegraphics[width=.6\textwidth]{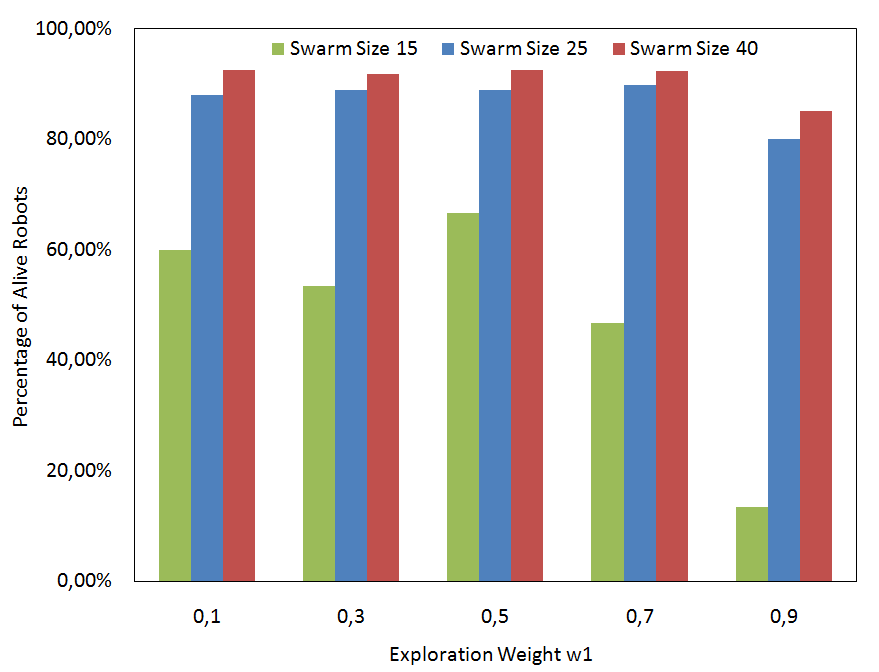}} \\
	\subfloat[][\emph]
	{\includegraphics[width=.6\textwidth]{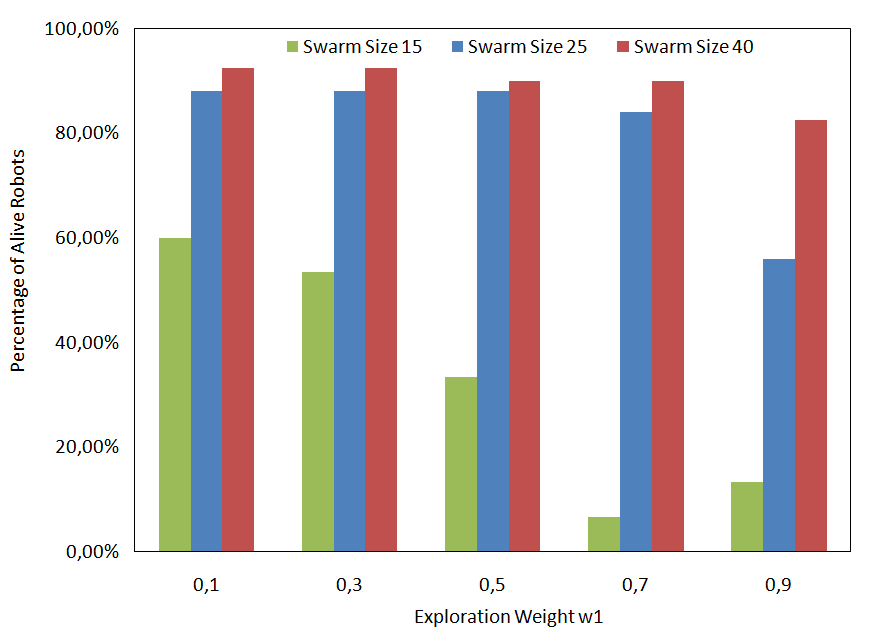}} \quad
	\caption{Percentage of alive robots in a 50x50 grid area, varying the dimension of the swarm of robots and 3 robots needed to handle a target. (a) 10 targets  (b) 15 targets  (c) 20 targets  }
	\label{fig:Dynamic50x50Alive}
\end{figure}

The percentages of alive robots, evaluated considering different sizes of areas, swarm size and dispersed targets, are summarized in Fig.~\ref{fig:Dynamic50x50Alive}. The figure illustrates that for a small robots team, if a greater importance is given to the exploration task, some reduction of the alive robots is obtained. This behaviour seems to be influenced by the number of dispersed targets. However, increasing the swarm size leads to no significant differences. This can be justified by previous motivations; if $w_1$ is high, the robots may be less likely to respond to the help requests, thus leading to the coordinator robots be trapped into targets's locations waiting for others to arrive and consequently increasing the probability of some explosions. Therefore, unbalanced resources can cause severe resource wastage, lead to the potential failure of the robots due to both the energy limitation and potential explosion risks.

In almost all experiments, the performance fluctuates according to the number of disseminated targets. This may indicate that the solution, would be greatly influenced by the recruiting weight value.

\newpage
\section{Conclusions}

In this paper, we have investigated a bi-objective optimization problem for robot coordination and exploration tasks. We have considered the exploration and manipulation of hazardous targets such as mines by a swarm of robots in search and rescue mission. Specifically, we have modeled the problem as a bi-objective model and  the weighted sum method is used to find a trade-off between the two tasks by varying different weighted values. The proposed strategy is bio-inspired and proves to be robust, effective with the use of limited resources in a balanced way.

In order to test the validity of our proposed model and the influence of the weight values, a Java-based simulator has been developed and implemented.
Different experimental scenarios have been considered to suitably evaluate the impact of the weight values on the critical parameters of the problem such as the dimension of the area, number of disseminated targets and number of robots to coordinate.
The results have demonstrated that the choice of the right compromise between the two tasks is not straightforward.  Generally speaking, the proper values depend on the application context. In most cases, the trade-off between the two objectives is highly correlated with the number of targets dispersed in the area, compared to the dimension of robot swarm. In these cases, more importance should be attributed to $w_2$ so that ($w_2$ $\geq$ 0.5).
However, in general case, balanced weights $w_1$ and $w_2$ (around 0.5) can offer a better trade-off.

Possible future works include the extension of methods to dynamically adjust the weights during the mission so as to be adaptive to the resource of the robots or other constraints. Furthermore, it will be useful to consider more realistic models and practical issues that robots may face in real-world scenario such as sensing and communication failure. In addition, the proposed method can be modified to potentially deal with the unknown but mobile targets in an unknown area, which will be explored further in future work.


\bibliographystyle{plain}

\bibliography{mybibfile2}

\end{document}